\documentclass[letterpaper, 10 pt, journal, twoside]{IEEEtran}

\ifCLASSINFOpdf
\else
\fi
\usepackage[utf8]{inputenc}
\usepackage[T1]{fontenc}
\usepackage{mathrsfs}
\usepackage{amsmath}
\usepackage{amssymb}
\usepackage{amsfonts}
\usepackage{amstext}
\usepackage{pgfplots}
\usepackage{graphicx}

\usepackage{tabstackengine}
\usepackage{mathtools}
\usepackage{multirow}
\usepackage{subcaption}
\usepackage{cite}
\usepackage{hyperref}
\usepackage{cleveref}
\graphicspath{ {./images/} }
\usepackage{geometry}
\usepackage{afterpage}

\begin{document}
\newgeometry{
letterpaper,
left=49pt,
right=49pt,
top=57pt,
bottom=48pt
}
\afterpage{\aftergroup\restoregeometry}
\title{Occlusion-aware Risk Assessment and \\ Driving Strategy for Autonomous Vehicles \\Using Simplified Reachability Quantification}

% \author{Michael~Shell,~\IEEEmembership{Member,~IEEE,}
%         John~Doe,~\IEEEmembership{Fellow,~OSA,}
%         and~Jane~Doe,~\IEEEmembership{Life~Fellow,~IEEE}% <-this % stops a space
% \thanks{M. Shell was with the Department
% of Electrical and Computer Engineering, Georgia Institute of Technology, Atlanta,
% GA, 30332 USA e-mail: (see http://www.michaelshell.org/contact.html).}% <-this % stops a space
% \thanks{J. Doe and J. Doe are with Anonymous University.}% <-this % stops a space
% \thanks{Manuscript received April 19, 2005; revised August 26, 2015.}}
\author{Hyunwoo Park$^{1}$, Jongseo Choi$^{1}$, Hyuntai Chin$^{1}$, Sang-Hyun Lee$^{2}$ and Doosan Baek$^{1,2*}$%
\thanks{Manuscript received: JULY, 31, 2023; Accepted OCTOBER, 13, 2023.}%Use only for final RAL version
\thanks{This paper was recommended for publication by Editor Aniket Bera upon evaluation of the Associate Editor and Reviewers' comments.}
\thanks{*This work was supported by the Korea Agency for Infrastructure Technology Advancement (KAIA) grant funded by the Ministry of Land, Infrastructure and Transport (RS-2021-KA160853, Road traffic Infrastructure monitoring and emergency recovery support service technology development).}% <-this % stops a space
    \thanks{$^1$ ThorDrive, Seoul, 07268, Republic of Korea}%
    \thanks{{\{\tt\small hwpark, jschoi, htchin, dsbaek\}@thordrive.ai}}%
    \thanks{$^2$ Seoul National University, Seoul, Republic of Korea}%
    \thanks{{\tt\small slee01@snu.ac.kr}}%
    \thanks{$^*$ Corresponding author}%
    \thanks{Digital Object Identifier (DOI): see top of this page.}
}

\markboth{IEEE ROBOTICS AND AUTOMATION LETTERS. PREPRINT VERSION. ACCEPTED OCTOBER, 2023}
{Park \MakeLowercase{\textit{et al.}}: Occlusion Risk Assessment and driving strategy}




\maketitle

\begin{abstract}
One of the unresolved challenges for autonomous vehicles is safe navigation among occluded pedestrians and vehicles. Previous approaches included generating phantom vehicles and assessing their risk, but they often made the ego vehicle overly conservative or could not conduct a real-time risk assessment in heavily occluded situations. We propose an efficient occlusion-aware risk assessment method using \emph{simplified reachability quantification} that quantifies the reachability of phantom agents with a simple distribution model on phantom agents' state. Furthermore, we propose a driving strategy for safe and efficient navigation in occluded areas that sets the speed limit of an autonomous vehicle using the risk of phantom agents. Simulations were conducted to evaluate the performance of the proposed method in various occlusion scenarios involving other vehicles and obstacles. Compared with the baseline case of no occlusion-aware risk assessment, the proposed method increased the traversal time of an intersection by 1.48 times but decreased the average collision rate and discomfort score by up to 6.14 times and 5.03 times, respectively. The proposed method has shown the state-of-the-art level of time efficiency with constant time complexity and computational time of less than 5 ms.
\end{abstract}

% \begin{IEEEkeywords}
% IEEE, IEEEtran, journal, \LaTeX, paper, template.
% \end{IEEEkeywords}

% \fi
\IEEEpeerreviewmaketitle

\section{Introduction}
\begin{figure}[t]
    \centering
    \includegraphics[width=\columnwidth]{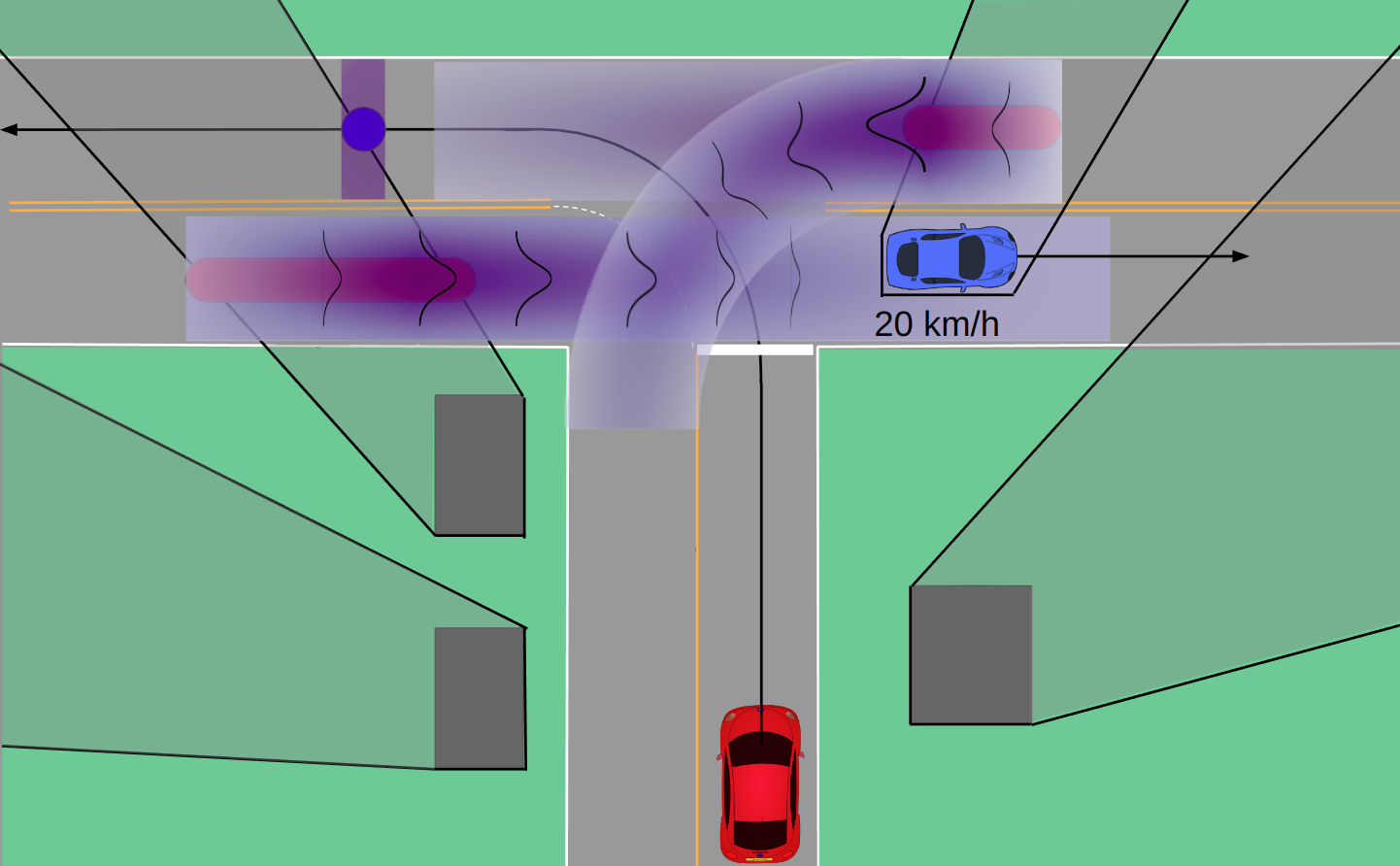}
    % \captionsetup{size=small}
    \captionsetup{size=footnotesize}
    \caption{Ego vehicle (red) entering an intersection. Occlusions due to obstacles (dark gray) and another vehicle (blue) are shaded in gray. Positions of potential hidden vehicles that can collide with the ego vehicle are indicated by red lines and blue point. The blue point has same direction with the ego vehicle, while the red lines do not. The risk along the centerline of the road due to potential hidden vehicles is represented as purple.
    The darker the purple gets, the riskier it is. Furthermore, the risk is proportional to a normal distribution where the random variable is the distance from the centerline of the road.}
    \label{fig:main}
\end{figure}

% \section{INTRODUCTION}
% Before correction
% LiDARs, cameras, and radars are often used in autonomous vehicles and other mobile robots to perceive surroundings. But due to their nature of line-of-sight trait, they have a critical disadvantage under occlusion in common situations like Fig.\ref{fig:main}. When occlusion is present in the scene, there could be potentially dangerous situations where traffic participants who were in the occluded area may come out from it and move into the driving corridor \cite{raksincharoensak2016motion, gilroy2019overcoming}. Human drivers could safely drive under occlusion by decreasing velocity depending on the extent of occlusion and its riskiness, yet not losing efficiency. Autonomous vehicles must also assess the potential risks due to occlusion and its extent of risk. Previous works solved this problem by generating hypothetical agents called Phantom Agent(PA) in the occluded region. In \cite{koschi2020set}, they proposed a set-based prediction that contains every possible future behavior of PAs. However, they didn't quantify the occlusion risk, which could make the ego vehicle overly conservative. In \cite{yu2020risk,yu2019occlusion}, they quantify the occlusion risk by sampling the particles which represent PAs in the occluded area. However, in heavily occluded areas, sampling the particles of the occluded area gets impossible to keep in real-time due to intensive computation.
 \IEEEPARstart{A}{utonomous} vehicles and other mobile robots often use LiDAR, cameras, and radar to perceive their surroundings. However, these sensors generally only work according to the line-of-sight, which gives them a critical disadvantage in common occluded situations as shown in Fig. \ref{fig:main}. A scenario with occluded areas can lead to potentially dangerous scenarios where other vehicles or pedestrians may suddenly appear in the route of the autonomous vehicle \cite{raksincharoensak2016motion, gilroy2019overcoming}. Human drivers handle occlusion by decreasing velocity sufficiently to lower the risk without losing efficiency. For autonomous vehicles, a common approach to occlusion-aware risk assessment has been to generate phantom agents (PAs) in occluded areas. In \cite{koschi2020set}, they proposed a set-based prediction method that considers every possible future behavior of PAs. However, they did not quantify the occlusion risk, which can make the ego vehicle behave overly conservatively. In \cite{yu2020risk}, and \cite{yu2019occlusion}, they quantified the occlusion risk by sampling particles that represent PAs in the occluded area. However, such an approach can lead to intensive computation in heavily occluded areas and make the risk assessment impossible to perform in real-time.
 
% In recent years, autonomous vehicles are sharing roads with human drivers. Naturally, autonomous vehicles are encountering various situations where their limited sensor range dangers the other human drivers and themselves\cite{raksincharoensak2016motion, gilroy2019overcoming}. To ensure safety, autonomous vehicles must overcome their limited sensing capabilities. 

 % Previous works \cite{yu2019occlusion,yu2020risk,lee2017collision,koschi2020set,orzechowski2018tackling} have tackled this problem by introducing "phantom/hidden" agents in occlusion area.
 % \cite{nager2019lies} have successfully managed to consider the hidden agents reasoning over time, yet they lack generality and efficiency. \cite{koschi2020set} have used a concept called Responsibility-Sensitive Safety(RSS)[13,!([1-5] in set-based paper) ] to verify safety. However driving strategy that keeps RSS is inefficient and could even cause an accident that they don't have a responsibility.!(More survey needed about RSS causing an accident). 

In this paper, we propose an efficient method for occlusion-aware risk assessment and a planning strategy for safe navigation through occluded areas. The risk assessment is based on \emph{simplified reachability quantification} (SRQ) that quantifies the ability of the PAs to reach a certain position using a simple distribution model on the PAs' state. The model-based quantification derived constant time complexity of the proposed method. The planning strategy sets speed limit in a position where occlusion risk is presented. This strategy is effective and can be easily integrated with any other planning and control algorithms because it plans a safe motion of the ego vehicle using only speed limit.

The main contributions are as follows:
\begin{itemize}

\item The proposed method achieves constant time complexity and assesses heavily occluded scenarios in real time using SRQ.

% \item The proposed method is robust in various driving environments and can consider different traffic participants, including pedestrians.

\item The proposed method can handle various types of driving environment, such as occluded vehicles in an intersection and occluded pedestrians behind street parking cars.

\item The proposed method achieves remarkable decreases in ride discomfort and the collision rate compared to the baseline case of no occlusion-aware risk assessment.

\item The proposed method is easily integrated with any other planning and control algorithms.

% We validate our algorithm in various types of occluded scenarios in the CARLA simulator, Our method resulted in a remarkable decrease in ride discomfort and collision rate.
% \item Our method is easily integrated with any planning and control algorithms.

% Before 0626, Exact date unknown
% \item Our method is generally applicable in arbitrary traffic situations quantifying the occlusion risk.

% \item By using the \emph{Simplified Reachability Quantification} to assess the occlusion risk of phantom vehicles, computational efficiency has enormously increased.

% \item Our algorithm is capable of assessing the risk of phantom pedestrians in cases like street parking scenarios, and jaywalking pedestrians.

% \item The proposed driving strategy is easily integrated with any planning and control algorithms.

\end{itemize}

The remainder of the paper is organized as follows: Section \ref{related} reviews related works of occlusion risk assessment and driving strategy utilizing it. Section \ref{preliminaries} defines problem setting and primary concepts, which would be used in later sections. Section \ref{method} describes how our method finds PAs that are relevant to the ego vehicle and assesses their risks. Then, motion planning using the occlusion-aware risk assessment is proposed. Section \ref{evaluation} shows how our method is evaluated in the CARLA simulator\cite{dosovitskiy2017carla} and the real world. Section \ref{result} analyzes the evaluation results and shows how the key metrics are improved. Section \ref{conclusion_and_future_work} concludes the proposed method and discusses future works.\\


\section{Related work} \label{related}

% Before correction
% Most of the previous works could be classified by following two categories: \emph{Probabilistic based methods} \cite{yu2019occlusion,yu2020risk,mcgill2019probabilistic,lee2017collision,wang2020generating}, and \emph{Over-approximation methods} \cite{orzechowski2018tackling,koschi2020set,nager2019lies}. There are also other kinds of methods variated from two methods, recreated itself by using concepts like \emph{Sequential Reasoning}\cite{nager2019lies, sanchez2022foresee}. On the following paragraphs, a brief explanation of occlusion risk assessment methods, its limitation, and driving strategies using them will be described by category.
Previous approaches to occlusion-aware risk assessment can be divided into two categories, namely \emph{probabilistic based methods} \cite{yu2019occlusion,yu2020risk,mcgill2019probabilistic,lee2017collision,wang2020generating} and \emph{over-approximation methods} \cite{orzechowski2018tackling,koschi2020set,nager2019lies}. There are also other methods variated from two methods and used concepts, such as \emph{Sequential Reasoning}\cite{nager2019lies, sanchez2022foresee}. These categories are briefly reviewed here.

% \emph{Driving Strategy} of above methods are different by the occlusion risk assessment method they use.
% Before Correction
% {\bf Probabilistic Methods}: Probabilistic methods assess an occlusion risk by formulating the problem probabilistically. They quantify the occlusion risk of each point. In \cite{yu2019occlusion}, they sampled \emph{the potential vehicles}' predicted positions to assess occlusion risk however they didn't consider the type of the \emph{the potential vehicles} to filter out the unnecessary ones, and the particles during the prediction horizon. Sampling each and every particle was also computationally inefficient. Whereas our approach covers mentioned limits and efficiently generates a distribution of \emph{the potential vehicles} within the prediction horizon. In \cite{yu2020risk}, they calculated $\mathbf{BRS}$(Backward Reachability Set) of every particle of $\mathbf{FRS}$(Forward Reachability Set) of the ego-vehicle to assess occlusion risk which could be inefficient when the number of the particle gets large. Also, they consider every possible control input of the \emph{potential vehicles} which is computationally inefficient since most of the vehicles drive along the lane.
Probabilistic methods quantify the occlusion risk by formulating the problem probabilistically. In \cite{yu2019occlusion}, they sampled potential positions of the hidden \emph{potential vehicles} (phantom vehicles) to assess the risk. However, they did not consider the types of the phantom vehicles to filter out unnecessary ones, and sampling the particles(phantom vehicles) was inefficient. In \cite{yu2020risk}, they calculated backward reachability set ($\mathbf{BRS}$) of every particle of forward reachability set ($\mathbf{FRS}$) of the ego vehicle to assess the occlusion risk. This method can be inefficient particularly with a large number of particles in heavily occluded areas. In addition, they considered every possible control input of the phantom vehicles, which is computationally inefficient because most vehicles have few options when driving in the real world.

Over-approximation methods are a special case of probabilistic method\cite{yu2019occlusion} that consider the probability of every phantom vehicle as 100\%. In \cite{koschi2020set}, they used an \emph{edge}, first introduced by \cite{orzechowski2018tackling}, to classify \emph{static} and \emph{dynamic} phantom vehicles. The classification results were then used to assess the occlusion risk.  Over-approximation methods are efficient because they don't quantify the risk, however they tend to make the ego vehicle behave conservatively and even freeze in some corner cases where the ego vehicle must take a risk to pass through. 

% In \cite{koschi2020set,orzechowski2018tackling,wang2020generating}, 
%doesn't represent the risk-taking attitude of the expert driver nor the-ego vehicle can't move in some edge-cases.

Studies that used \emph{sequential reasoning} considered only necessary agents based on observations over time. In \cite{nager2019lies}, they used an \emph{over-approximation method} to track hidden agents conservatively over time and formulate a passive safety ($\emph{p-safe}$) \cite{macek2009towards} planning strategy using \emph{braking inevitable collision states} \cite{bouraine2012provably} and \emph{responsibility sensitive safety} \cite{shalev2017formal}. However, they tracked every possible agent within their sensor range, which is inefficient for heavily occluded areas and with long-range sensors.

% {\bf Active Perception}:
% % \cite{zhang2021safe, narksri2022occlusion,wang2020generating}
% Active perception is a strategy that intentionally changes the vehicle's state according to sensing strategy\cite{bajcsy1988active}. In occluded situations, it could be interpreted as a strategy that increases the visibility for safety and efficient behavior. In \cite{wang2020generating}, they penalized the trajectories that have low visibility. % cons of this visibility method 
% In \cite{zhang2021safe}, they accounted the vehicle's visibility by using A* search with a receding horizon method that minimizes the intersection between \emph{Danger zone} and \emph{Hidden zone}.

Driving strategies differ depending on the approach used for occlusion-aware risk assessment. Among probabilistic methods, \cite{yu2019occlusion} used optimization to generate a trajectory with a low occlusion risk, \cite{wang2020generating} used a minimum cost function including an occlusion risk cost to select a trajectory, and \cite{mcgill2019probabilistic} reduced the velocity of the ego vehicle when the risk was sufficiently high before it entered an intersection. Among over-approximation methods, \cite{koschi2020set,orzechowski2018tackling} obtained fail-safe trajectory \cite{pek2018computationally,althoff2016set} that did not collide with any other phantom vehicles.\\
% Add and review this paper! \cite{naumann2019safe}, \cite{yu2020risk

% {\bf} dynamic occlusion
% {\bf} Pedestrian

% {\bf POMDP}: In recent motion planning, Partially Observable Markov Decision Process (POMDP) have optimized the behavior of AVs, thus reducing the collision risk caused by occlusion25–28. However, the calculation amount of the POMDP method increases exponentially with the number of states required for operation, which limits its \\
% 1)S. Brechtel, T. Gindele, and R. Dillmann, “Solving continuous pomdps: Value iteration with incremental learning of an efficient space representation,” in Inter- national Conference on Machine Learning, 2013.\\
% 2)[16] M. Bouton, A. Nakhaei, K. Fujimura, and M. J. Kochen- derfer, “Scalable decision making with sensor occlu- sions for autonomous driving,” in IEEE International Conference on Robotics and Automation (ICRA), 2018.\\
% 3) [17] S. Russell and A. L. Zimdars, “Q-decomposition for reinforcement learning agents,” in Proceedings of the 20th International Conference on International Confer- ence on Machine Learning, ser. ICML’03, Washington, DC, USA: AAAI Press, 2003, pp. 656–663.\\
% 4) • Pedestrian collision avoidance system for scenarios with occlusions / 2019
% IV cited by 25  POMDP Paper
% 1.Pedestrian collision avoidance system for scenarios with occlusions

% {\bf Road Network} : \cite{bender2014lanelets,koschi2020set,koschi2017spot}

%\IEEEpubidadjcol

\section{Preliminaries} \label{preliminaries}
In this section, problem setting and primary concepts used in later sections will be defined. Let $t \in \mathbb{R}$ be the time, $x(t)$ be the state of the system at time $t$ 
in the state space \raisebox{0.4ex}{$\chi$}, and $u$ be the ego vehicle's control input in the action space $\mathbb{U}$. The environment surrounding the ego vehicle comprises $n \in \mathbb{Z}$ number of lanes $l \in L $. Each lane $l$ has a set of \emph{continuous} and \emph{sequential} centerline points denoted by $P^{l}$. The $k$-th lane $l_k$, where ($0 \leq k \leq n \ , \ k \in \mathbb{Z}$), has $m$ centerline points $P^{l_k} = \{p^{l_k}_1, p^{l_k}_2, p^{l_k}_3, \cdots, p^{l_k}_m\}$ in the Cartesian space $p \in P \subset \mathbb{R}^2$ and road width of $l^{k}_w$. The route of the ego vehicle is predefined as points following certain lanes $l \in L$ to reach the goal in the Cartesian space, which is represented as $route(x(t_0)) = \{x(t_0), x(t_1), x(t_2), \cdots , x(t_k)\}$ where $x(t) \in \raisebox{0.4ex}{$\chi$}$.\\[4pt]
{\bf Definition 1. (Observable Polygon, $\mathbb{O}$)}. $\mathbb{O}$ is generated by
the line-of-sight sensors from the ego vehicle and comprises points $p$ in the Cartesian space $p \in \mathbb{O} \subset \mathbb{R}^2$.\\[4pt]
{\bf Definition 2. (Phantom Agent}, PA). PA is a potential hidden agent outside the observable polygon and is modeled as a point mass that allows small agents like cyclists and children could be included. PAs can be further classified as phantom vehicles (PVs) or phantom pedestrians (PPs).

PVs are assumed to always be on the road and drive along the lane. They can have various velocities and even exceed the speed limit of the road. In contrast, PPs can be anywhere because they are not physically constrained unlike cars on the sidewalk. In \cite{nager2019lies}, they also considered \emph{illegally-behaving pedestrians} that walk on roads.\\[4pt]
%  Given initial state $x(t_0)$ at time $t_0$ and fixed time horizon $T_{fix}$, the every possible set of states of a vehicle $x(t)$ could be represent using the action space of a vehicle $\mathbb{U}$ as follows:\\
% \begin{equation}
% \begin{split}
%     f(x_0,T_{fix}) := \{& x(t) \in \raisebox{0.4ex}{$\chi$} | \dot{x}(t) = u(t), \forall u(t) \in \mathbb{U},\\
%     &\forall t \in [t_0,t_0 + T_{fix}] \}
%     % f(x_0,T) := \{ x(t) \in \raisebox{0.4ex}{$\chi$} | \Dot{x_t} =  u(t)\}
% \end{split}
% \end{equation}
{\bf Definition 3. (Forward Reachable Set, $FRS$)} Assume the vehicle's dynamics are described as following ordinary differential equation:
\begin{equation}
\begin{split}
    \dot{x}(t) = f(x(t),u(t))
\end{split}
\end{equation}
The FRS is a set of states that could be reached in given an initial state $x_0$ at time $t_0$ and fixed time horizon $\text{T}_\text{fix}$: 
\begin{equation}
\begin{split}
    FRS(x_0,\text{T}_\text{fix}) := \{& x(t) \in \raisebox{0.4ex}{$\chi$} | \dot{x}(t) = f(x(t),u(t)),\\ &\forall u(t) \in \mathbb{U},
    \forall t \in [t_0,t_0 + \text{T}_\text{fix}] \}
\end{split}
\end{equation}
{\bf Definition 4. (Backward Reachable Set, $BRS$)} \\
The BRS is a set of initial states $x_0$ at time $t_0$ that can reach the given final state $x_{f}$ within the fixed time horizon $\text{T}_\text{fix}$:

\begin{equation}
\begin{split}
    BRS(x_f,\text{T}_\text{fix}):= \{ x_0 \in \raisebox{0.4ex}{$\chi$} | x_f \in FRS(x_0,\text{T}_\text{fix}), \forall u(t) \in \mathbb{U}, \\
    \dot{x}(t) = f(x(t),u(t)), \forall t \in [t_0,t_0 + \text{T}_\text{fix}] \}
    % &\forall t \in [t_0,t_0 + \text{T}_\text{fix}] \} \\
\end{split}
\end{equation}

\section{Method} \label{method}
% Before correction
% We first generate PAs in the occluded region. This step consists of \emph{Node Classification}(\ref{sec:Node_Classification}) and \emph{Inferring Phantom Agent Zone}(\ref{sec:Phantom_Vehicle_Zone}). \emph{Node Classification} outlines the positions of PAs using road situations and occluded area, \emph{Inferring Phantom Agent Zone} finds every position of PAs that have the ability to affect the ego vehicle. Second, we define occlusion risk by quantifying the risk of the PAs using \emph{Simplified Reachability Quantification}(\ref{main:risk_assessment}). Third, using the occlusion risk we derived, we develop a driving strategy by setting a speed limit for the ego vehicle where occlusion risk presents.\\
The proposed method has three steps. First, PAs in occluded areas are generated by \emph{node classification} (Section \ref{sec:Node_Classification}) and \emph{inferring the phantom agent zone} (Section \ref{sec:Phantom_Vehicle_Zone}). Then, an approach we call \emph{Simplified Reachability Quantification} (Section \ref{sec:SRQ}) that quantifies the reachability of the PVs is proposed. Next, the occlusion risks of PAs are defined by SRQ. Finally, a driving strategy that sets a speed limit depending on the occlusion risk is developed.

\begin{figure*}[t!]
    \centering
    % 0.235
    \begin{subfigure}[t]{0.24\textwidth}        %% or \columnwidth
        \centering
        \includegraphics[width=\linewidth]{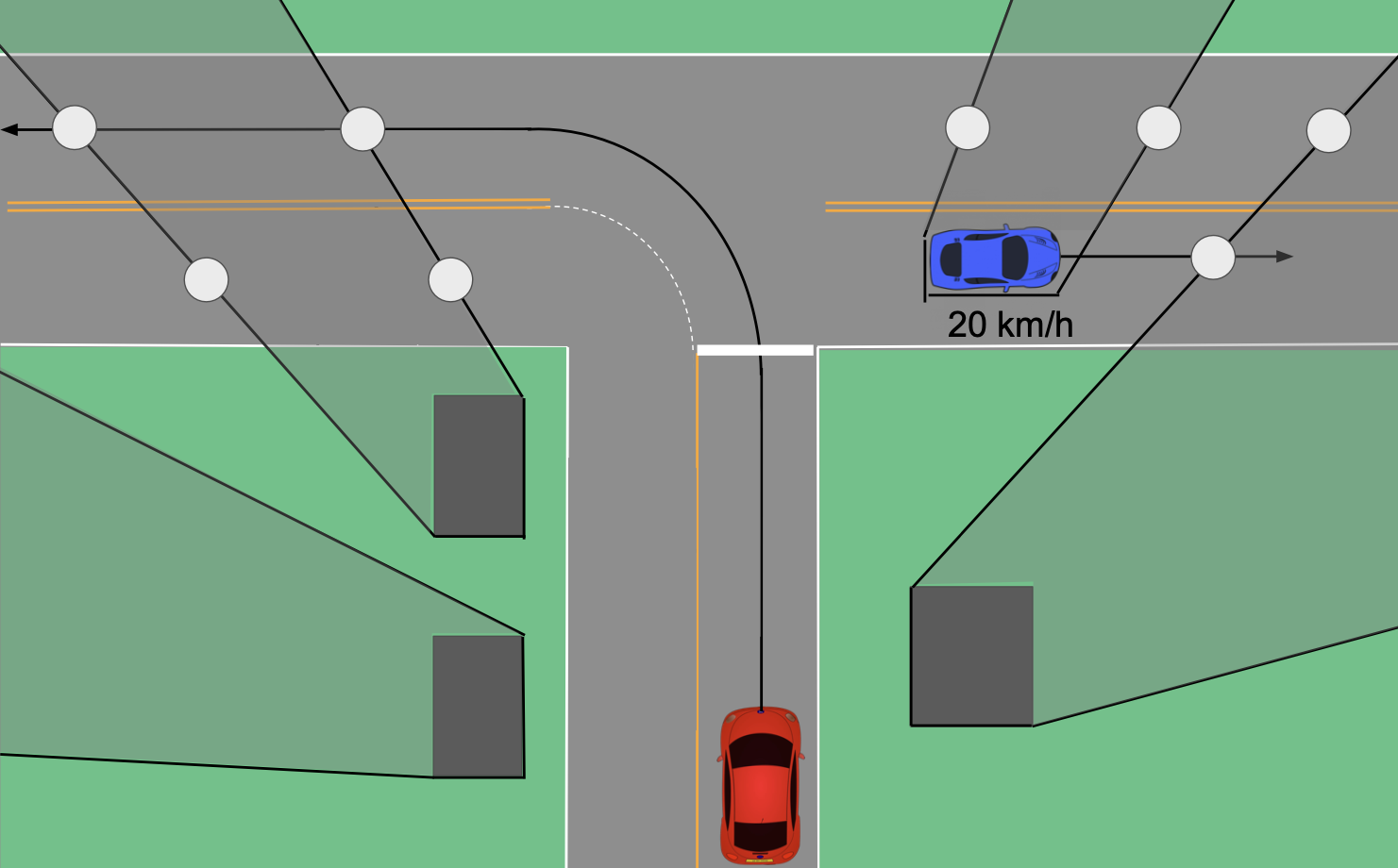}
        \caption{}
        \label{fig:main_a}
    \end{subfigure}
    \begin{subfigure}[t]{0.24\textwidth}        %% or \columnwidth
        \centering
        \includegraphics[width=\linewidth]{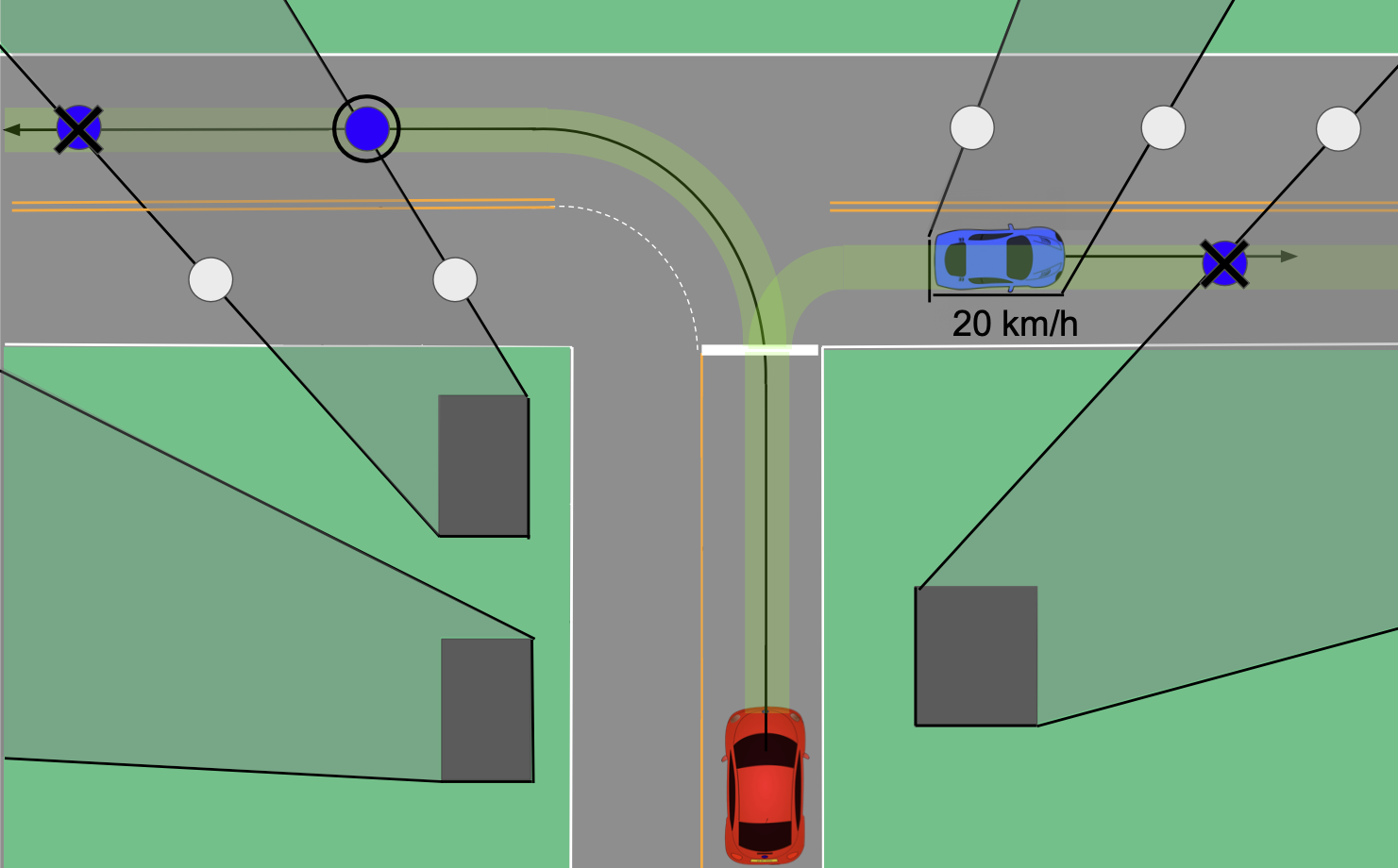}
        \caption{}
        \label{fig:main_b}
    \end{subfigure}
    \begin{subfigure}[t]{0.24\textwidth}        %% or \columnwidth
        \centering
        \includegraphics[width=\linewidth]{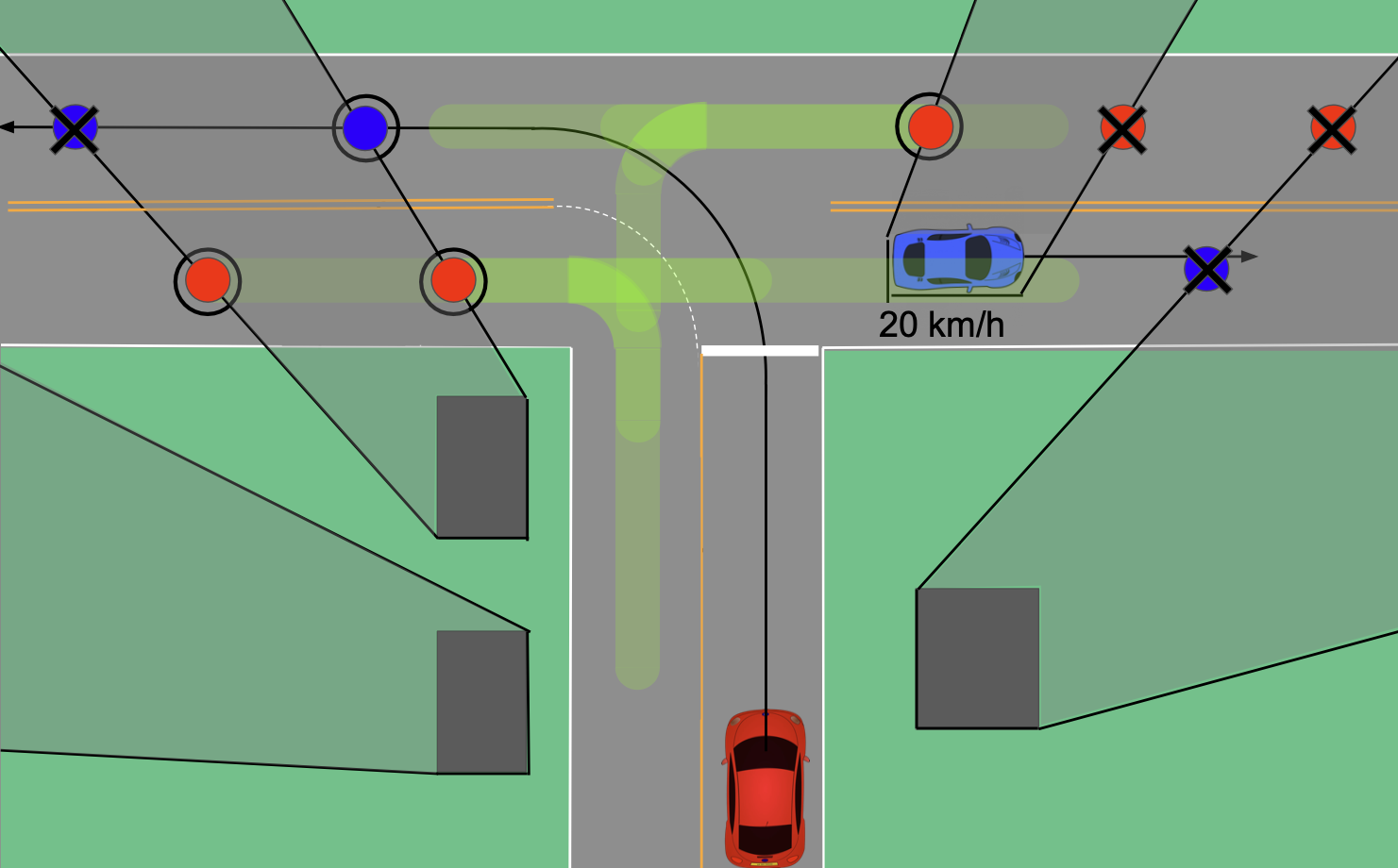}
        \caption{}
        \label{fig:main_c}
    \end{subfigure}
    \begin{subfigure}[t]{0.24\textwidth}        %% or \columnwidth
        \centering
        \includegraphics[width=\linewidth]{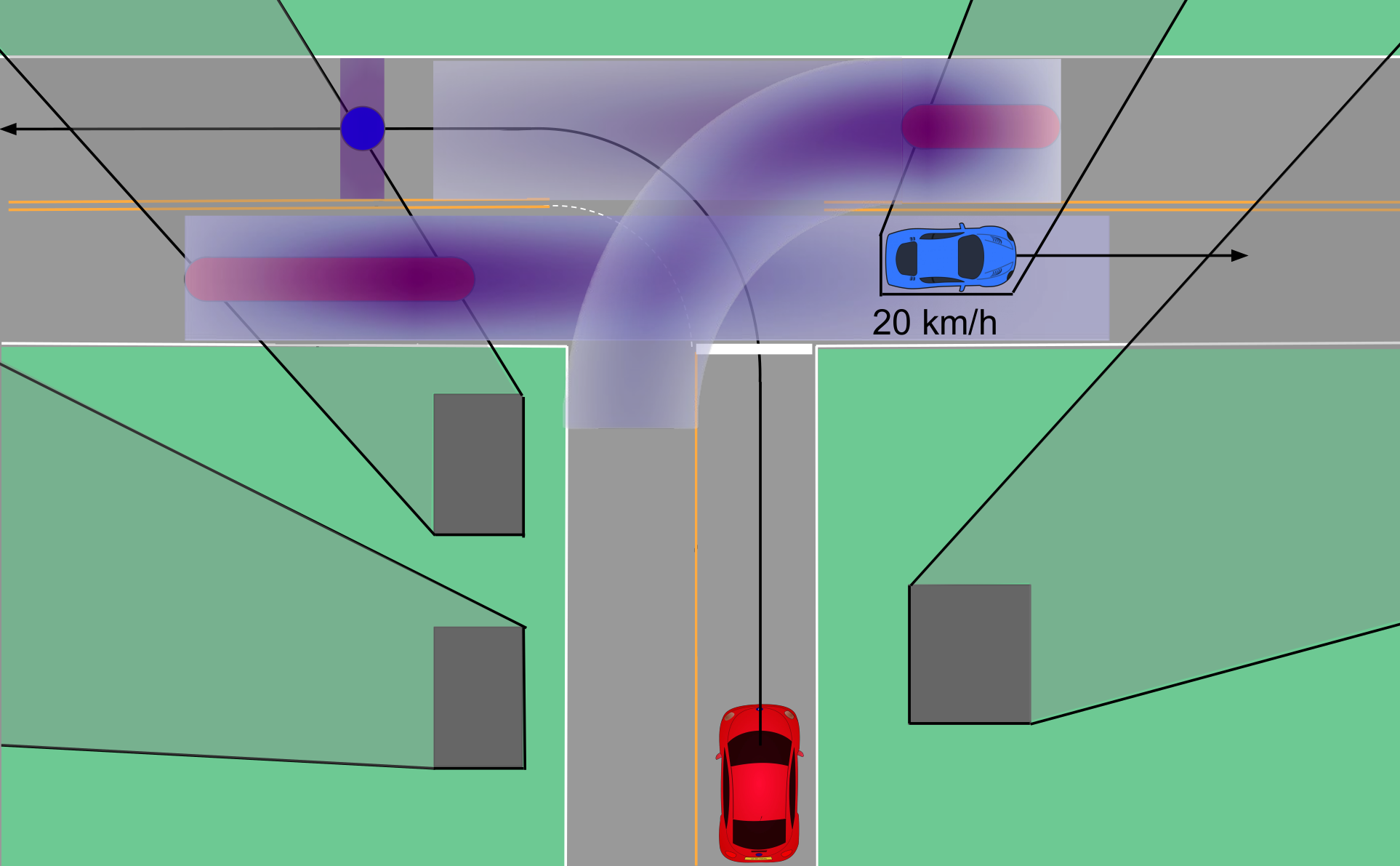}
        \caption{}
        \label{fig:main_d}
    \end{subfigure}
    \captionsetup{size=footnotesize}
    \caption{Node classification: (a) application scenario where an ego vehicle (red) navigates an intersection having occluded areas (shaded in gray) and another vehicle (blue). Intersecting nodes are represented as white points. (b) Relevant static node (circled blue point) and irrelevant static nodes (crossed-out blue points) are classified according to the FRS of the dynamic nodes (shaded in yellow). (c) Relevant dynamic nodes (circled red points) and irrelevant dynamic nodes (crossed-out red points) are classified according to the FRS of the dynamic nodes (shaded in yellow). (d) Obtained phantom vehicle zone and its occlusion risk. The magenta carries the same meaning as in Fig. \ref{fig:main}}
    \label{fig:node_classification}
\end{figure*}

\subsection{Node Classification} \label{sec:Node_Classification}
% Before correction
% The potential positions of PVs are points on the centerlines of lanes $P^{l_{k}}$ in the occluded area. However, considering every PV is inefficient. We could efficiently assess risk by classifying PV, whether the PV would \emph{statically} or \emph{dynamically} risk the ego vehicle\cite{orzechowski2018tackling}, and by removing PVs that can't reach the route of the ego vehicle in the prediction horizon $\text{T}_\text{pred}$. The above efficiency is achieved, by a sequence of procedures consisting of \emph{Node Classification} and \emph{inferring the phantom agent zone}. \emph{Node Classification} is a cornerstone of the procedure to achieve efficiency. The result of it is used for the \emph{inferring the phantom agent zone} by a search starting from each node. The illustration of \emph{Node Classification} is in Fig.\ref{fig:node_classification} (a)-(c) and it consists as follows:
Node classification is based on the edge classification which classifies edges into static, dynamic, relevant, and irrelevant edges \cite{koschi2020set}. They over-approximate the shape of the PVs and represented PVs as edges. Static edges are PVs that have same direction with the ego vehicle, whereas dynamic edges are PVs in different direction than that of the ego vehicle. Relevant edges are PVs that can reach the route of the ego vehicle in the prediction horizon $\text{T}_\text{pred}$, while irrelevant edges are PVs that cannot reach the route of the ego vehicle. In \cite{koschi2020set}, they also over-approximate the PVs' positions to be the closest position to the ego vehicle. In our method, node is introduced instead of the edge to obtain PV intervals and to avoid the over-approximations of the PVs.

Node classification is a sequence of procedures used to improve the efficiency of the risk assessment by classifying PVs according to whether they pose a \emph{static} or \emph{dynamic} risk to the ego vehicle and by removing irrelevant PVs. Figs. \ref{fig:node_classification}a-c show the three main steps of node classification.

\subsubsection{Intersection of Observable Polygon and Lanes}
% Before correction
% First, \emph{intersected Nodes} $\mathbb{I}$ is defined as intersection of $\mathbb{O}$ and centerline of lanes $P^{l_k}$. 
% \emph{Intersected Nodes} are used for finding out potential positions of PVs and we only need the centerline of lanes $P^{l_k}$ to assess the risk of every PV. (Detailed explanations are in \ref{risk_assment:static_phantom_vehicle}, \ref{risk_assment:dynamic_phantom_vehicle}). However, we exclude the intersected points which overlap with other vehicles or are too close to other vehicles. Since the positions of PVs inferred intersected points will be so small or overlap with objects that even small agents can't exist. 
\emph{Intersecting nodes} $\mathbb{I}$ are defined as the intersection of the observable polygon $\mathbb{O}$ and the centerline of lanes $P^{l_k}$. 
% Detailed explanations are in \ref{sec:Phantom_Vehicle_Zone}.
\begin{equation}
    \mathbb{I} = \{ p \in \mathbb{R}^2 | p = \mathbb{O} \cap P^{l_k}, k \in \{1,2,3, \cdots m\} \}
\end{equation}

$\mathbb{I}$ are used to find the potential positions of PVs. $\mathbb{I}$ that overlap with or are too close to other vehicles that even small agents cannot exist are excluded.
\subsubsection{Static/Dynamic Node Classification}
% Before Correction
% Second, we classify $\mathbb{I}$ into static node $\mathbb{S}$ and dynamic node $\mathbb{D}$. The static/dynamic node came from static/dynamic edges in \cite{koschi2020set}. They named it static edge since they over-approximate PVs and the riskiest move the PV generated from static edge could have is stopping in front of PV. However we modified it since we don't over-approximate the shape of the PV and we need an interval to find every possible position of PVs, not over-approximating by using an edge.
% We first classify the $\mathbb{I}$ as $\mathbb{S}$, if $\mathbb{I}$ has any intersection with $FRS(x_{ego},\text{T}_\text{fix})$. Then the rest of the $\mathbb{I}$ is $\mathbb{D}$. Static nodes can be visited by the ego vehicle in a given fixed time horizon and dynamic nodes can't be visited. This property makes inferring positions of PVs and risk assessment efficient.
The intersecting nodes $\mathbb{I}$ are then classified as static nodes $\mathbb{S}$ or dynamic nodes $\mathbb{D}$. $\mathbb{I}$ are classified as $\mathbb{S}$ if it has any intersection with $FRS(x_{ego},\text{T}_\text{pred})$, and the remainder of $\mathbb{I}$ are classified as $\mathbb{D}$:
\begin{equation} \label{eq::StaticDynamicNode}
\begin{split}
    \mathbb{S} = FRS(x_{ego},\text{T}_\text{pred}) \cap \mathbb{I},
    \quad \mathbb{D} = \mathbb{I} - \mathbb{S}\\
\end{split}
\end{equation}

$\mathbb{S}$ can be visited by the ego vehicle in the given time horizon, but $\mathbb{D}$ cannot. Note
that the concept of static and dynamic are used to characterize other
terms in the following sections.
\subsubsection{Relevant/Irrelevant Node Classification}
% Now we know whether a node \emph{statically} or \emph{dynamically} risks the ego vehicle. 
% Before Correction
% Before inferring the positions of the PVs using static/dynamic nodes, we have to 
% classify whether the node is necessary or not. There are unnecessary nodes that don't have to be considered because their FRS doesn't collide with \emph{route path} of the ego vehicle in $\text{T}_\text{pred}$. We define a node as a \emph{Relevant Node} $\mathbb{RN}$ if $FRS(i,\text{T}_\text{pred}) \ where \ i \in \mathbb{I}$ collides with the route path of the ego vehicle and as \emph{Irrelevant Node} $\mathbb{IN}$ if it doesn't. However if the $i \in \mathbb{S}$, we only consider the first $\mathbb{S}$ that is reached by the route path of the ego vehicle as a \emph{relevant} like \cite{koschi2020set} did. Since \emph{static} occlusion risk by other $\mathbb{S}$ are already taken care of by the first one. Note that the concept \emph{relevant} will be used to characterize other terms as a trait in the following sections.
Next, the static nodes $\mathbb{S}$ and dynamic nodes $\mathbb{D}$ need to be classified as \emph{relevant} or \emph{irrelevant} nodes. Irrelevant nodes do not need to be considered because their FRS does not collide with the route of the ego vehicle within $\text{T}_\text{pred}$. A node is classified as relevant ($\mathbb{RN}$) if $FRS(i,\text{T}_\text{pred}), \ where \ i \in \mathbb{I}$, collides with the route of the ego vehicle and as irrelevant ($\mathbb{IN}$) if it does not:
% \cite{orzechowski2018tackling} 
\begin{equation} \label{eq::RelevantIrrelvantNode}
\begin{split}
    \mathbb{RN} = \{i \in \mathbb{I} | FRS(i,\text{T}_\text{pred}) \cap route(x(t_0)) \neq \varnothing \}\\
    \mathbb{IN} = \{i \in \mathbb{I} | FRS(i,\text{T}_\text{pred}) \cap route(x(t_0)) = \varnothing \}
\end{split}
\end{equation}

However, if $i \in \mathbb{S}$, then only the first $\mathbb{S}$ reached by the route of the ego vehicle is considered relevant\cite{koschi2020set} because the risk of the other $\mathbb{S}$ is accounted for by the first one. Note that the concept of relevance is used to characterize other terms in the following sections.

\subsection{Inferring the phantom agent zone} \label{sec:Phantom_Vehicle_Zone}
The \emph{phantom agent zone} is a set of occluded points where \emph{relevant} PAs exist. It comprises the \emph{phantom vehicle zone} (PVZ) and \emph{phantom pedestrian zone} (PPZ). The PVZ comprises the \emph{phantom vehicle set} (PVS), which is a set of occluded points in a lane that are continuously connected to each other. We assume that one \emph{relevant} PV exists in every PVS, therefore every PVS must include at least one \emph{relevant node} $\mathbb{RN}$. If a PVS includes a \emph{relevant} $\mathbb{D}$, it is defined as a \emph{dynamic} PVS. However, if it includes \emph{relevant} $\mathbb{S}$, then only the \emph{relevant} $\mathbb{S}$ is defined as the \emph{static} PVS. In Fig. \ref{fig:node_classification}d, the dynamic PVS is represented as a red line, while the static PVS is represented as blue dot.

% For every point in the occluded area $p \in \mathbb{O}^c$ and every lane $l_k$, the PVZ is defined as follows:
% \begin{equation} \label{eq::phatnom_vehicle_zone}
% \begin{split}
%     &\mathbb{PVZ} := \{ p \in P^{l_k} | FRS(p,\text{T}_\text{pred}) \cap route(x(t_0)) \neq \varnothing\}\\
% \end{split}
% \end{equation}

% In this section, \emph{Phantom Vehicle Zone} and \emph{Phantom Pedestrian Zone}, which would be used for risk assessment later, will be defined. First, \emph{Phantom Vehicle Zone} will be defined using \emph{Relevant Nodes} to find longitudinal position along the lane of PVs that actually risks the ego vehicle. Every PVS is defined using each \emph{Relevant Nodes} and whether the \emph{Relevant Node} is \emph{static} or \emph{dynamic} decides the PVS to be Dynamic PVS or Static PVS. Second, \emph{Phantom Pedestrian Zone} will be defined without using the result of \emph{Node Classification}.
% Phantom vehicle zone $\mathbb{VZ}$ is a set of occluded points on the centerline of the lane and it is \emph{longitudinal} positions of PVs that could risk the ego vehicle in $\text{T}_\text{pred}$. $\mathbb{VZ}$ could be divided to \emph{Phantom Vehicle Set}(PVS) that is set of points that are attached to each other. Every PVS includes at least one $\mathbb{I}$. If a PVS includes $\mathbb{D}$ it is named \emph{Dynamic Phantom Vehicle Set}(Dynamic PVS), if it includes $\mathbb{S}$ it is named \emph{Static Phantom Vehicle Set}(Static PVS).
% In Fig.\ref{fig:main}, Dynamic PVS is represented as red line and Static PVS is represented as blue dot.
\subsubsection{Static Phantom Vehicle Set}
% Defining positions of every PV that \emph{statically} risk the ego vehicle as Static PVS is dangerous and inefficient. If we found every PV and assess its risk(details are in \ref{main:risk_assessment}), it would be riskier since occlusion risk would be scattered. Also, we observed that static occlusion risk is often removed before it affects the ego vehicle. This characteristic of static occlusion risk makes defining every PV as Static PVS unnecessary and also prevents the ego vehicle from conservative movement even if we define \emph{Relevant} $\mathbb{S}$ as Static PVS which is the most dangerous assumption for the ego vehicle. Therefore defining Static PVS as \emph{Relevant} $\mathbb{S}$ is an efficient and safest way to consider the positions of \emph{Static} PVs.
Defining the static PVS as encompassing the positions of every PV that poses a static risk to the ego vehicle is dangerous and inefficient because the occlusion risk would be scattered. Furthermore, we have observed that the static occlusion risk is often removed before it affects the ego vehicle. Therefore, the most efficient and safest way to consider the positions of static PVs is to define the static PVS as encompassing the positions of relevant $\mathbb{S}$.

% (1) Even though we found Static PVS and assess its risk, it's risk is often removed before while driving along the route path. 
% (2) Inefficient, not safe
% 1)  
% This can't be reason why I don't find every Static PVS.
% (2) If the route is feasible yet the route is occluded, it is likely occluded by other vehicle that is in front of the ego vehicle which isn't occlusion risk assessment.
\subsubsection{Dynamic Phantom Vehicle Set} 
\begin{figure}[t!]
    \centering
    \includegraphics[width =0.7\linewidth]{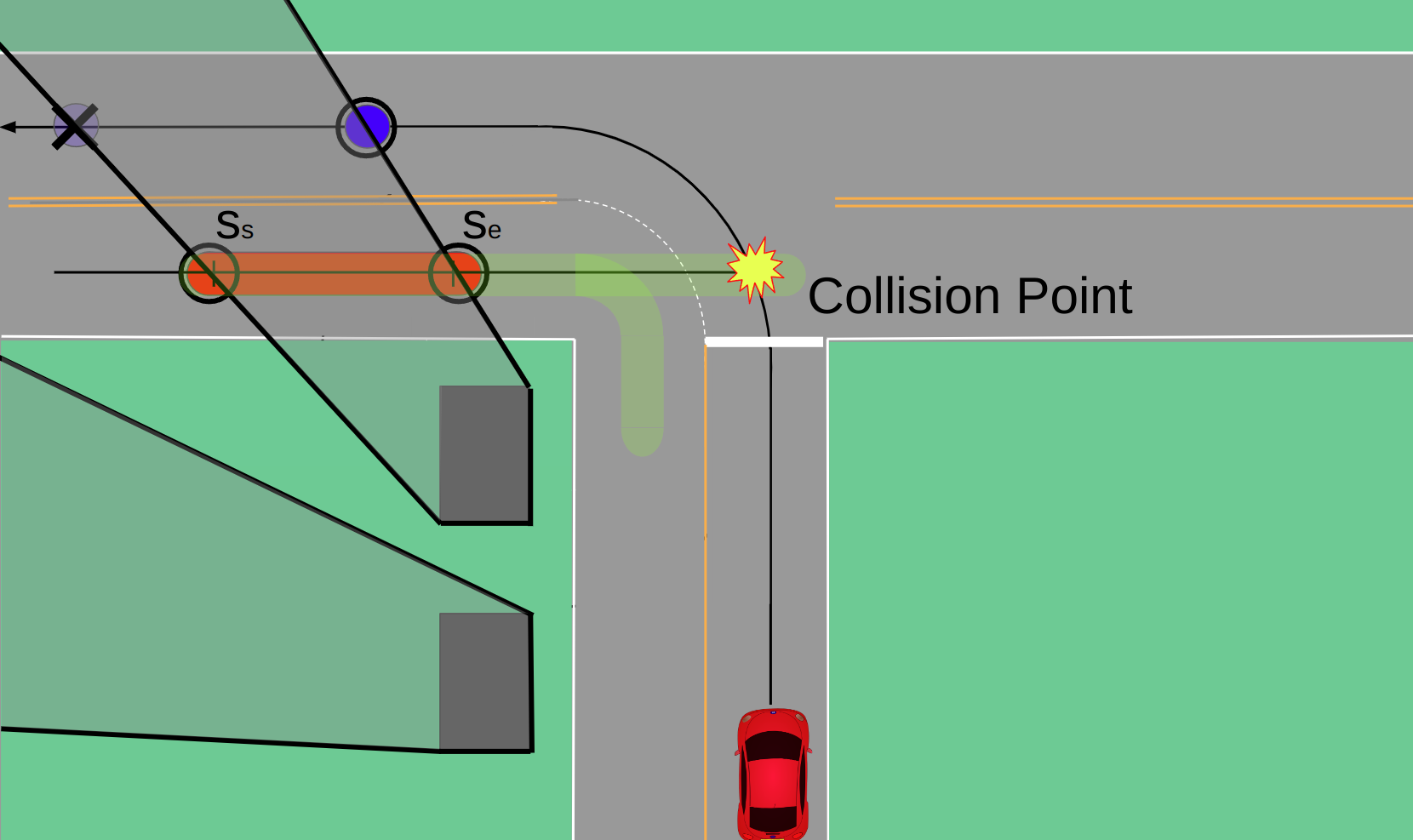}
    \captionsetup{size=footnotesize}
    \caption{Dynamic phantom vehicle set (PVS): The starting position of the dynamic PVS (red) is $(s_{s},0)$, and the end position is $(s_{e},0)$ where both positions are in the Frenet frame of PV. The forward reachable set (FRS) of the collision point is shaded in yellow.}
    \label{fig:example}
\end{figure}

% Before Correction
% Dynamic PVS could be found by following steps: (1) The starting point of Dynamic PVS is \emph{Dynamic Relevant Node}. (2) In the reverse direction of the lane, new point $p_{new}$ is added to Dynamic PVS from starting point until (3),(4) occurs. (3) If $p_{new}$ gets in to the $\mathbb{O}$, it stops. (4) If $FRS(p_{new},\text{T}_\text{pred})$ doesn't collide with the route path of the ego vehicle, it stops. Dynamic PVS in a common intersection scenario is shown in Fig.\ref{fig:example}. Using Dynamic PVS, Every potential \emph{dynamic} PV that is \emph{relevant} with the ego vehicle is considered.
The dynamic PVS is obtained as follows. First, the dynamic relevant nodes are taken as the starting point of the dynamic PVS. Then, new points $p_{new}$ are added to the dynamic PVS in the reverse direction of the lane until two conditions are satisfied: $p_{new}$ reaches $\mathbb{O}$ and $FRS(p_{new},\text{T}_\text{pred})$ does not collide with the route of the ego vehicle. Fig. \ref{fig:example} shows the dynamic PVS obtained for a common intersection scenario. The dynamic PVS considers every potential dynamic PV that is \emph{relevant} to the ego vehicle.

\subsubsection{Phantom Pedestrian Zone}
Unlike PVs, PPs can be anywhere. In \cite{nager2019lies}, they called PPs outside a sidewalk or crosswalk, \emph{illegally-behaving pedestrians}. Because the area of illegally-behaving pedestrians covers the area of legally-behaving pedestrians, we only considered illegally-behaving pedestrians. Given $\text{T}_\text{pred}$ and route of the ego vehicle at time $t_0$, the PPZ is defined as follows:
\begin{equation} \label{eq::phatnom_vehicle_zone}
\begin{split}
    PPZ :=  \{& p \in \mathbb{R}^2 | p = \mathbb{O}^c \cap BRS(x_k,\text{T}_\text{pred}),\\
    &x_k \in route(x(t_0)) \}
\end{split}
\end{equation}

\begin{figure}[t!]
    \centering
    \includegraphics[width = 0.7\linewidth]{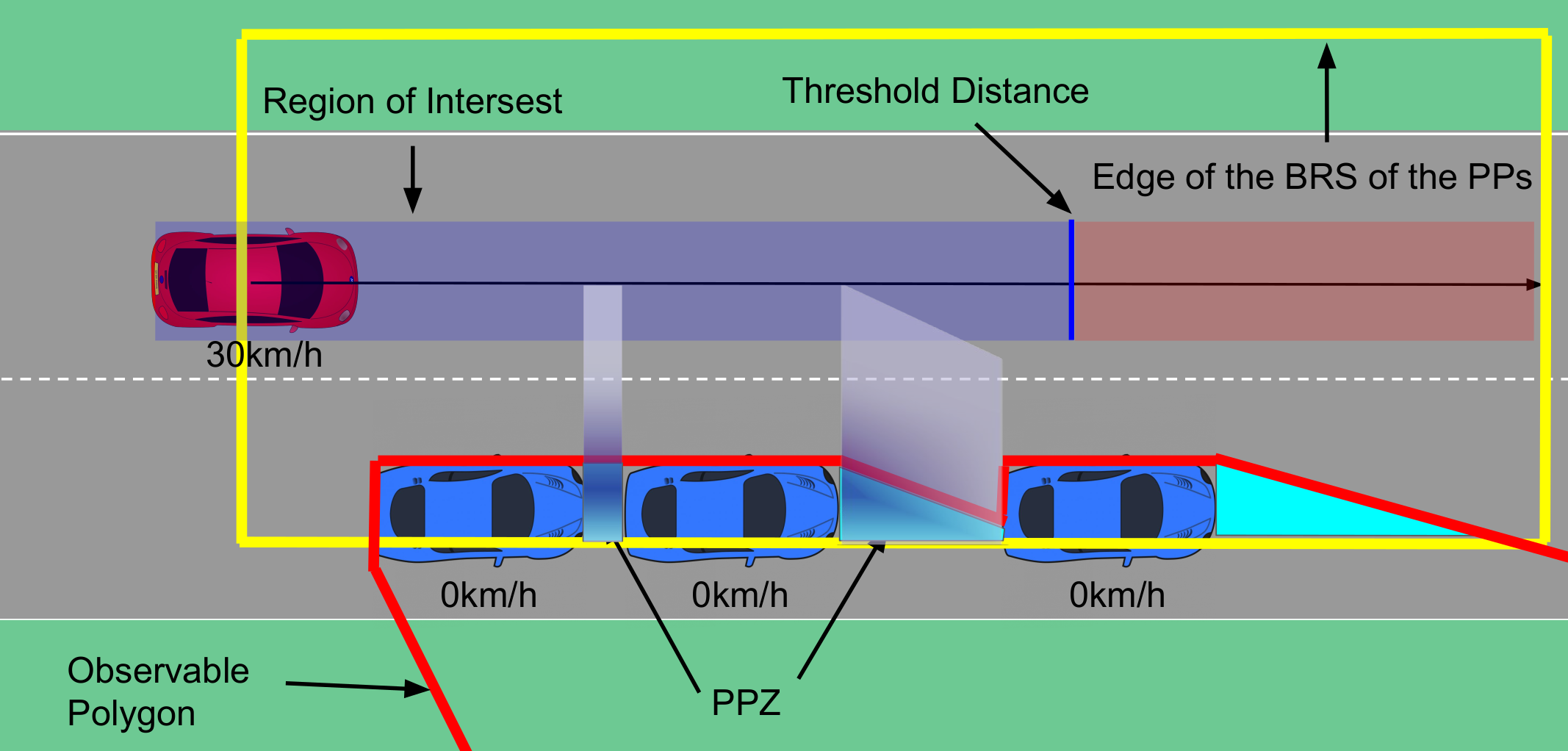}
    \captionsetup{size=footnotesize}
    \caption{Phantom pedestrian zone (PPZ): The ego vehicle (red) drives along other vehicles (blue) parked on the street. The PPZ (cyan) is derived from the BRS (yellow rectangle) of the route (black) and the observable polygon (red). The risk is represented as purple, and the darker the purple gets the riskier it is. The risk of PPs beyond a threshold distance (blue line) is filtered out.}
    \label{fig:pedestrian}
\end{figure}

% \begin{equation}
% PPZ := \{ BRS(x_k, \text{T}_\text{pred}) \cup \mathbb{O}^c \}
% \end{equation}
Unlike the PVZ, no nodes are available for obtaining the FRS of PP. Therefore, we use the BRS of the route of the ego vehicle to find the PPZ. The furthest PP (i.e., the edge of the BRS of the PPs) from the route path of the ego vehicle would be PP moving at maximum speed toward the route of the ego vehicle along the shortest path. For a road such as that shown in  Fig. \ref{fig:pedestrian}, the PPZ can simply be obtained as the rectangular area along the route from which the observable polygon is subtracted.

%\subsubsection{Time-series aware removal}
\subsection{Simplified Reachability Quantification} \label{sec:SRQ}

Strongly motivated by \cite{yu2019occlusion,yu2020risk}, we improved upon their occlusion-aware risk assessment method by reducing the computational load. Our approach, namely SRQ, quantifies the reachability of dynamic PVS using a simple distribution model as follows.

% In \cite{yu2019occlusion}, they ignore the PV in time horizon $t \in [0,\text{T}_\text{pred}]$, but only consider the state of the PV in time step $\text{T}_\text{pred}$ which is ignoring the PVs during $\text{T}_\text{pred}$. In \cite{yu2020risk}, they used backward reachability set(BRS) but they considered every control input of the PV which is inefficient since most of  the vehicles follows the lane.
% which isn't accurately sampling the PV.

SRQ quantifies the reachability of the dynamic PVS using the BRS with simple distribution model. Let $\text{T}_\text{pred}$ be the prediction horizon in which it is impossible to avoid a collision with a suddenly appearing vehicle due to the computational time of the autonomous driving stack, the maximum velocity of PV $\text{v}_\text{max}$, the dynamic PVS, $x$ be the initial position of a PV, $y$ be the final position of a PV, and $g(y)$ amount of element of $BRS(y,\text{T}_\text{pred})$, which represents the amount of PVs that can reach $y$ while satisfying below conditions. Here, PV is assumed to be driving along the lane. Therefore, $x$ and $y$ are defined as longitudinal positions in the Frenet frame of the PV, which can be anywhere within the dynamic PVS and can have random velocity. However, since no prior information is provided, the initial position of PV is assumed to be uniformly distributed along the DPVS $[s_s,s_e]$ and the velocity of PV is constant and uniformly distributed $[0,\text{v}_\text{max}]$ like \cite{yu2019occlusion} did. The function $g(y)$ is defined in three intervals according to the value of $y$: $I_1:=[s_{s},s_{e}]$, $I_2:=[s_{e},s_{s}+\text{v}_\text{max}\text{T}_\text{pred}]$, and $I_3:=[s_{s}+\text{v}_\text{max}\text{T}_\text{pred},s_{e}+\text{v}_\text{max}\text{T}_\text{pred}]$. Note that $s_{e}-s_{s} \leq \text{v}_\text{max}\text{T}_\text{pred}$ always satisfies by the definition of the dynamic PVS in Section \ref{sec:Phantom_Vehicle_Zone}. 

\begin{figure*}[!ht]
    \centering
    % 0.235
    \begin{subfigure}[t]{0.325\textwidth}        %% or \columnwidth
        \centering
        \includegraphics[width=\textwidth]{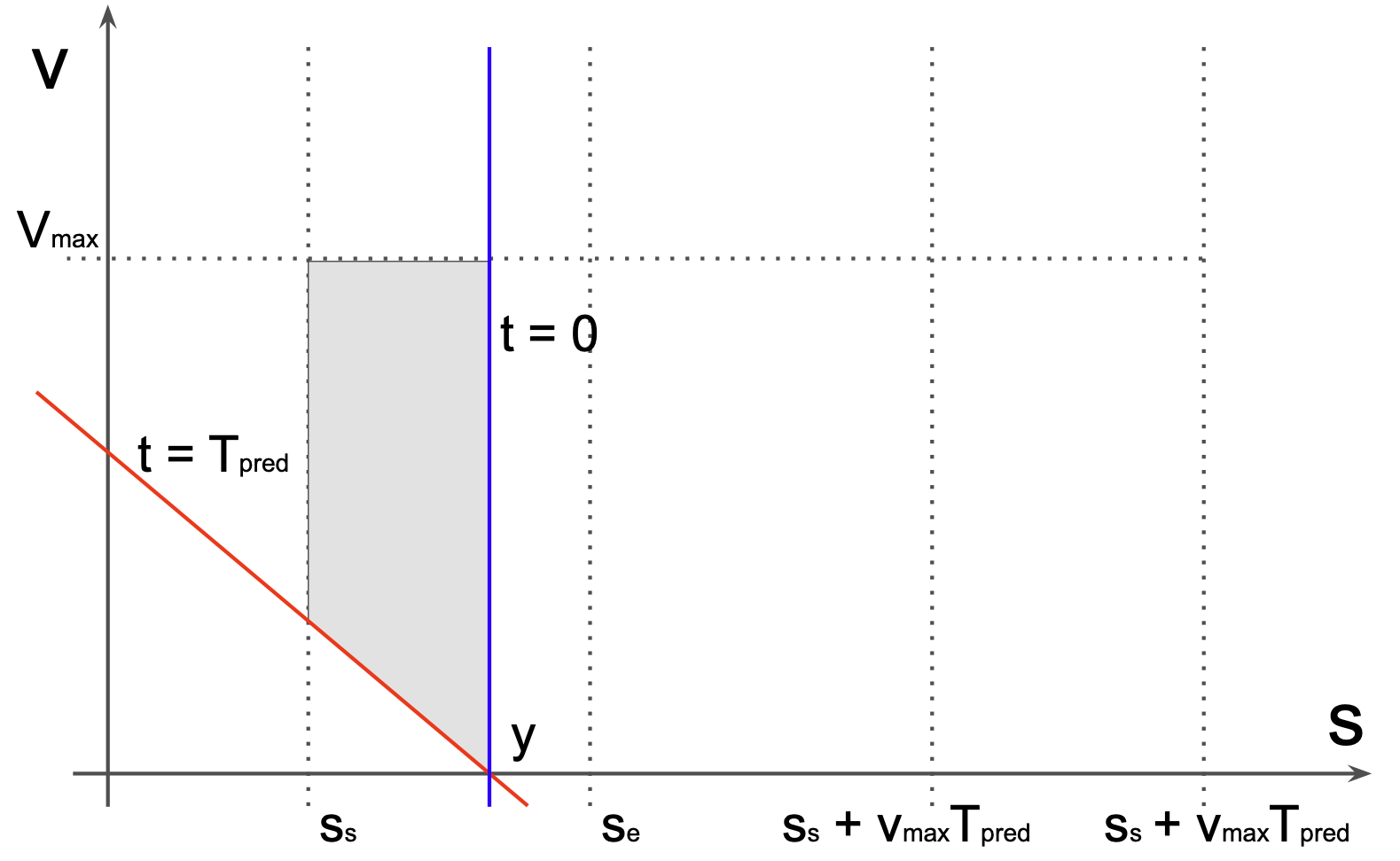}
        \label{fig:appendix1}
    \end{subfigure}
    \begin{subfigure}[t]{0.325\textwidth}        %% or \columnwidth
        \centering
        \includegraphics[width=\textwidth]{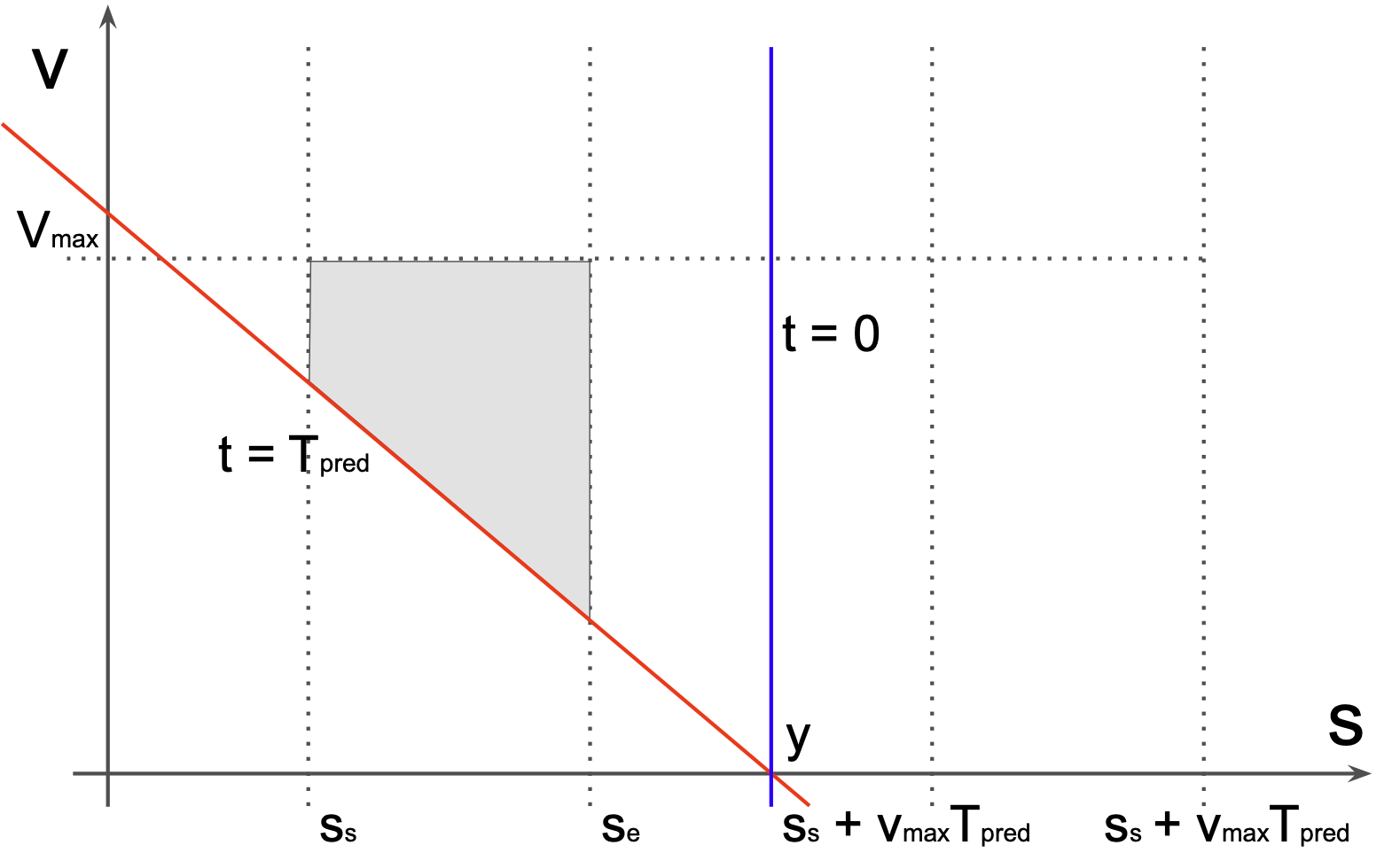}
        \label{fig:appendix2}
    \end{subfigure}
    \begin{subfigure}[t]{0.325\textwidth}        %% or \columnwidth
        \centering
        \includegraphics[width=\textwidth]{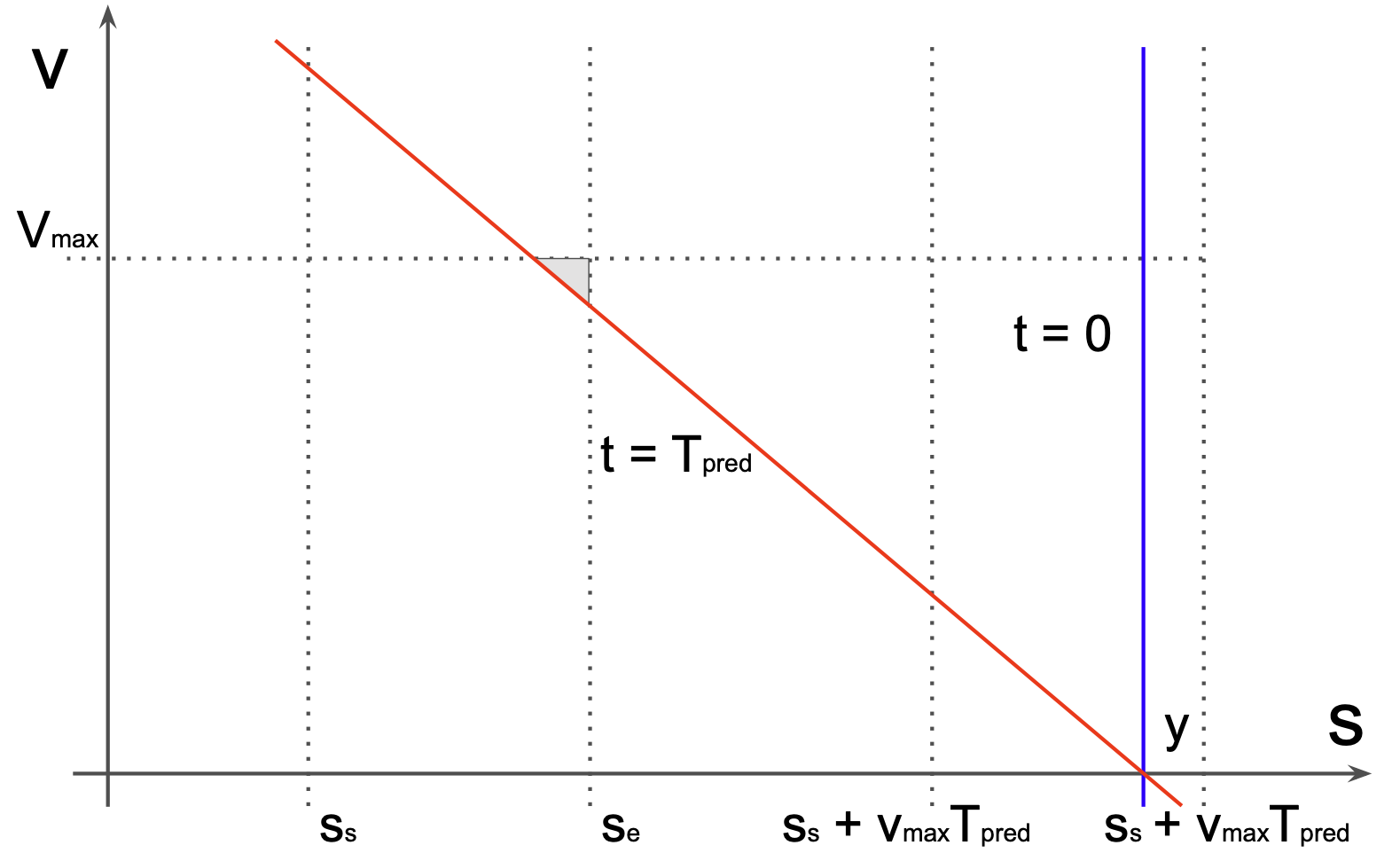}
        \label{fig:appendix3}
    \end{subfigure}
    \\[1ex]
    \captionsetup{size=footnotesize}
    \caption{The area of shaded in gray represents the $g(y)$. Each graph represents a different interval of $g(y)$ depending on the value of $y$.}
    \label{fig:appendix}
\end{figure*}

\begin{figure}[!t] 
    \centering
    \includegraphics[width = 0.7\linewidth]{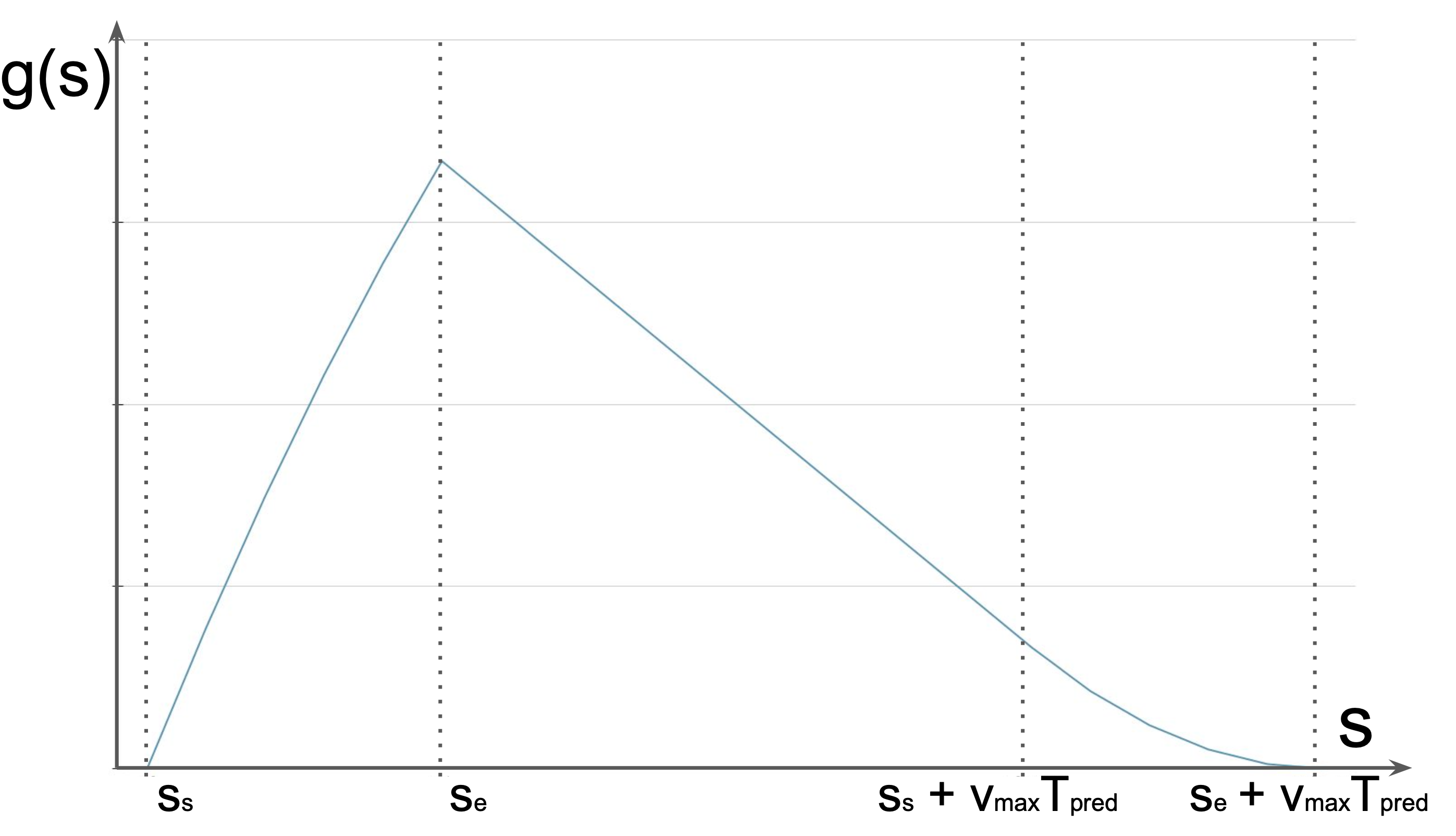}
    \captionsetup{size=footnotesize}
    \caption{SRQ along the longitudinal position in the Frenet frame of the dynamic PVS.}
    \label{fig:graph}
\end{figure}

Fig. \ref{fig:appendix} shows the BRS of PV depending on the value of $y$ that satisfies the aforementioned assumptions, as gray-shaded polygon. Because the initial position and velocity of PV are assumed to be uniformly distributed, the area of gray-shaded polygon represents the PV that can reach $y$. Note that the conditions $x \in [s_{s},s_{e}], v \in [0,\text{v}_\text{max}], t \in [0,\text{T}_\text{pred}]$ should be satisfied. The blue line represents the PV whose initial positions are already in $y$, therefore they reach $y$ at $t=0$ regardless of their velocities. The red line represents the PV that manages to reach $y$ at $t=\text{T}_\text{pred}$. Because the velocity of PV is assumed to be constant, $y = x + vt$ is always satisfied. Using this condition, $g(y)$ is easily derived. For convenience, $s$ is used instead of $y$ because $s$ is commonly used to represent the longitudinal position in the Frenet frame. This is graphed in Fig. \ref{fig:graph} and is derived as follows:\\[13pt]
$g(s)$\\
\begin{equation} \label{eq:final_equation_in_method}
\begin{split}
    :=
    \left\{\begin{array}{lll}
    {1 \over 2}(2\text{v}_\text{max}-{s-s_{s} \over \text{T}_\text{pred}})(s-s_{s}),&(s \in I_1)\\ [4pt]
    {1 \over 2}(2\text{v}_\text{max}-{s-s_{s} \over \text{T}_\text{pred}}-{s-s_{e} \over \text{T}_\text{pred}})(s_{e}-s_{s}),&(s \in I_2)\\ [4pt]
    {1 \over 2}(\text{v}_\text{max}-{s-s_{e} \over \text{T}_\text{pred}})(s_{e}-(s-\text{v}_\text{max}\text{T}_\text{pred})),&(s \in I_3)\\
    \end{array}\right.
\end{split}
\end{equation}

% Given $\text{T}_\text{pred}$ and Dynamic PVS, sampling size of velocity $N$, sampling size of positions $N$, we could calculate the ratio of the future positions $n/NP$ of PVs using the simplified bidirectional reachability model assuming velocity and initial positions are uniformly distributed. The start position of Dynamic PVS is $s_{s}$ and end position is $s_{e}$. See Fig.\ref{fig:example}.
% Let $x$ an initial position of a PV, $y$ a final position of a PV, the risk of the future positions in the Frenet frame of PVs is defined as follows: (Prove is in $\mathbf{APPENDIX.A-C})$)\\
Here, (\ref{eq:final_equation_in_method}) considers every possible motion of PV in every position of the dynamic PVS without prior information.

% \appendix 
% \section{A}
% \label{appendix::a}
% \begin{appendices}
% \section{} \label{appendix:a}

% \section*{APPENDIX}\label{appendix:a}
% \section{Mathematical Model}
% Here, we present the proof for the SRQ. The aim is to quantify the initial states of PV that can reach the collision point by using the BRS, which can be used to define the occlusion risk.

\subsection{Risk Assessment} \label{sec:risk_assessment}
% In this section, where the PVs and PPs would risk the ego vehicle and how risky it would be is found using \emph{Phantom Agent Zone}.
The PVZ and PPZ obtained from the previous steps is then applied to the risk assessment.

\subsubsection{Static Phantom Vehicle}\label{risk_assment:static_phantom_vehicle}
% Like we mentioned before, \emph{static} occlusion risk is often removed before the ego vehicle go near by.
% So it is inefficient to assume that Static PVS moves forward and it only makes the ego vehicle riskier. Due to above reasons we chose to consider Static PVS to stop in \emph{Relevant Static Node}.
For the same reason why static PVS is defined to encompass relevant $\mathbb{S}$, PV in the static PVS is assumed to be stopped, which is an most efficient and safest approach to assess its risk.

\subsubsection{Dynamic Phantom Vehicles}\label{risk_assment:dynamic_phantom_vehicle}

% If an element of the BRS of the collision point in Tpred is in an occluded area, then the PV poses an occlusion risk. In other words, the occluded area may contain a hidden vehicle with the ability to cause a collision.

$g(s)$ from SRQ represents the amount of PV that can reach $s$. It can be interpreted as the occluded area (i.e., dynamic PVS) may contain a hidden vehicle with the ability to reach the position $s$, and the probability to reach position $s$ increases as $g(s)$ increases. Therefore, $g(s)$ can be used to define an occlusion risk of dynamic PVS.

% If an element of the BRS of the collision point in Tpred is in an occluded area, then the PV poses an occlusion risk. In other words, the occluded area may contain a hidden vehicle with the ability to cause a collision. 

% However in (Section \ref{sec:SRQ}), PV is assumed to be uniformly distributed in the dynamic PVS. In other words, we assumed that an actual vehicle always exists in the dynamic PVS. 
In Section \ref{sec:Phantom_Vehicle_Zone}, we assumed that an actual vehicle always exists in the dynamic PVS. The probability that an actual vehicle exists in the dynamic PVS should also be considered. Naturally, the probability increases if the dynamic PVS is longer. Therefore, the occlusion risk should be defined to be increased with a longer dynamic PVS. The occlusion risk $o(s)$ considering the probability of an actual vehicle existing in the dynamic PVS is defined as follows:
\begin{equation} \label{eq:final_final_equation_in_method}
\begin{split}
    o(s) := (s_e - s_s) \cdot g(s)
\end{split}
\end{equation}

However, assessing the risk of a dynamic PVS of which the collision point is too far from the ego vehicle is unnecessary. A more efficient approach is to filter out the dynamic PVS of which the collision point is farther than \emph{static nodes} or a threshold distance, which is proportional to the current velocity of the ego vehicle.

Because not every vehicle drives along the centerline of a road, the lateral deviation of PVs should be considered, which we assume can be represented as a normal distribution of which confidence interval of lateral deviation $d$ (set to 90\% for evaluation) is $[-l^{k}_w/2, l^{k}_w/2]$. Let $w(d) := N(0,( {l^{k}_w \over 2 \times Z(1 - 0.5(1-d))})^2)$. Then, the final distribution of the risk of the dynamic PVS can be defined as follows:
\begin{equation}\label{eq:final_final_final_equation_in_method}
    r(s,d) := o(s) \times w(d)
\end{equation}
The occlusion risk that are on the route of the ego vehicle are illustrated in Fig. \ref{fig:node_classification}d.

% We assume that lateral deviation of \emph{Phantom Vehicle} 
\subsubsection{Phantom Pedestrians}
Unlike PVs, PPs can move in any direction. Thus, the FRS of a PP for a given $\text{T}_\text{pred}$ can be assumed as a circle. However, PPs moving away from the route of the ego vehicle do not have to be considered, therefore the FRS can be assumed as a semicircle toward the route. Without prior information, PP could be assumed to move straight toward the closest point along the route. Since the expected heading angle of PP would be straightforward to the closest point along the route.

 If the heading angle of the PP is fixed and with the same assumption used in risk assessment of PVs, the risk of PPs can be assessed in the same manner as for PV. Given $s_{s}, s_{e}, \text{T}_\text{pred}, \text{v}_\text{max}$, \eqref{eq:final_final_equation_in_method} can be used to assess the risk due to PPs. However, assessing the risk of PPs too far from the current position is unnecessary. Thus, such PPs should be filtered out as we did for unnecessary PVs as illustrated in Fig. \ref{fig:pedestrian}.

\subsection{Driving Strategy} \label{drivng_strategy}
% Before Correction
% In this section, a driving strategy that ensures the safety of the ego vehicle from occlusion risk is made using the result of \ref{main:risk_assessment}. There are lots of driving strategies for occlusion risk assessment. One could plan a trajectory that reduces an occlusion risk \cite{yu2019occlusion,yu2020risk,wang2020generating}, or plan the velocity of a fixed path \cite{sadou2004occlusions,naumann2019safe,mcgill2019probabilistic,bouton2018scalable}. We chose to plan the velocity of a fixed path. Since expert drivers rarely change a path(lateral deviation) in common occluded situations like intersections, roads with street parking cars, and narrow alleys. Also, parameter tuning of with/without occlusion risk for trajectory generation is hard or even won't work in the desired way.
There are lots of driving strategies when driving through an occluded area. One of them is to plan a trajectory that reduces the occlusion risk  \cite{yu2019occlusion,yu2020risk,wang2020generating} or plan the velocity profile of a fixed path \cite{sadou2004occlusions,naumann2019safe,mcgill2019probabilistic,bouton2018scalable}. The latter approach was chosen because expert drivers rarely change their path in common occluded scenarios such as intersections, roads with street parking, and narrow alleys. In addition, tuning the parameters for generating trajectories with and without occlusion risk would be difficult or would not work in the desired manner.

The occlusion risk obtained from SRQ can be used to control the velocity of the ego vehicle on a fixed path. The occlusion risk can be used directly as a \emph{cost function} for optimization or trajectory selection, lowering the \emph{target velocity}, or adding a \emph{speed limit} condition.

\begin{figure}[t]
    \centering
    \includegraphics[width = 0.8\linewidth]{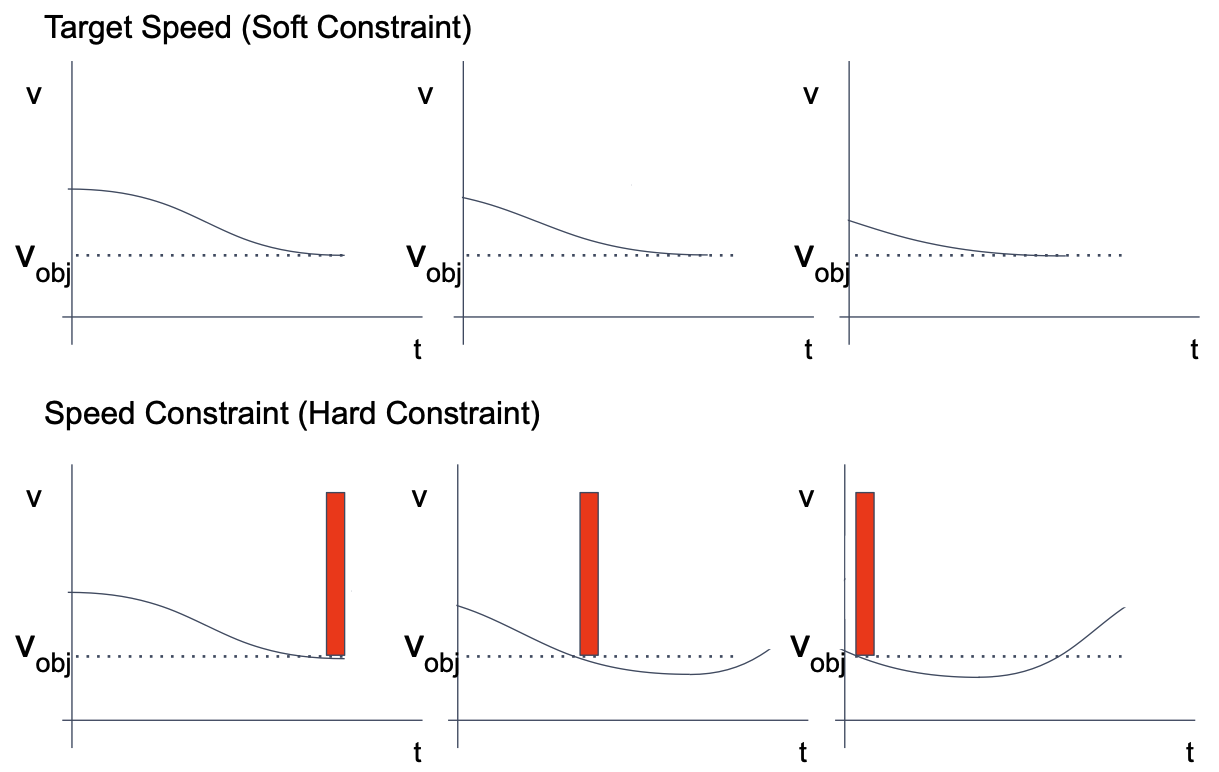}
    \captionsetup{size=footnotesize}
    \caption{Velocity planning delay. The same velocity is used as the target velocity and speed limit. The red square represents the speed limit due to occlusion risk.}
    \label{fig:speed_limit}
\end{figure}

Using the occlusion risk as cost for optimization can yield a high computational load or even be infeasible because the occlusion risk distribution is spread over a large area. Lowering the target velocity is ineffective if the velocity is planned for every time step, which will delay the time taken by the ego vehicle to reach the target velocity as shown in Fig. \ref{fig:speed_limit} \cite{werling2010optimal}. Adding a speed limit condition is an agile and effective approach that allows the ego vehicle to be aware of exact position where occlusion risk presents and guarantees that it slows down where expected. In addition, the speed limit is easily integrated with any planning algorithm by adding the extra constraint $v \leq v_{speed \textunderscore limit}$ or selecting a trajectory that satisfies the speed limit among sampled trajectories.
% Also lowering target velocity (2) can't be aware of exact position where the occlusion risks exist.
% Adding a speed limit (3) is an agile and effective approach that can be aware of an exact position and guarantees to slow down at the exact position. Also, it could be easily integrated with any planning algorithm by adding an extra constraint $v \leq v_{speed \textunderscore limit}$ in the optimization process or selecting a trajectory that satisfies the speed limit from sampled trajectories.
\subsubsection{Speed Limit}
 The speed limit $v^{occ}_{speed \textunderscore limit}$ and its position $p_{speed \textunderscore limit}$ due to the occlusion risk are obtained as follows. First, the points along the route of the ego vehicle where an occlusion risk exists should be determined. Second, these points should be clustered. Third, the \emph{weighted average} of the occlusion risk for these points should be obtained. Then, the speed limit and its position can be defined as follows:
% mathematical equations
\begin{equation} \label{eq:weighted_average}
\begin{split}
    p^{[c]}_{speed \textunderscore limit} := { \sum_{k = 1}^{N^{[c]}} { r^{[c]}_k \cdot p^{[c]}_k \over r^{[c]}_{total}} }\\
    % \sum_{n=1}^{\infty} 2^{-n} = 1
\end{split}
\end{equation}

where $c \in C$ is the cluster of points with an occlusion risk, $p_{speed \textunderscore limit}^{[c]}$ is the weighted average position of occlusion risk for each cluster $c$, $N^{[c]}$ is number of points in cluster $c$, $p^{[c]}_{k}$ is the position of the point, $r^{[c]}_{k}$ is the occlusion risk of the point, and $r^{[c]}_{total}$ is the total sum of the occlusion risk defined in (\ref{eq:final_final_final_equation_in_method}) for cluster $c$. Then,\\ [13pt]
% Weighted Average with Primitive Func
% $g(s_0 + s) :=$
$v^{occ}_{speed \textunderscore limit}$
\begin{equation} \label{eq:speed_limit}
\begin{split}
    =
    \left\{\begin{array}{ll}
    {\text{v}^{\text{occ}}_{\text{min}} - \text{v}^{\text{occ}}_{\text{max}} \over \text{c}^{\text{th}}_{\text{max}} - \text{c}^{\text{th}}_{\text{min}}}(r^{[c]}_{total}- \text{c}^{\text{th}}_{\text{min}}) 
    + \text{v}^{\text{occ}}_{\text{max}},
    & (\text{c}^{\text{th}}_{\text{min}} \leq r^{[c]}_{total} \leq \text{c}^{\text{th}}_{\text{max}})\\[6pt]
    \text{v}^{\text{occ}}_{\text{min}},&(\text{c}^{\text{th}}_{\text{max}} < r^{[c]}_{total})\
    % \text{v}^{\text{occ}}_{\text{min}},\qquad \qquad \qqaud \: \! \  \ (\text{c}^{\text{th}}_{\text{max}} < r^{[c]}_{total}\\
    \end{array}\right.
\end{split}
\end{equation}
% $v_{road \textunderscore speed \textunderscore limit}$ is the speed limit of the road

where $\text{c}^{\text{th}}_{\text{min}}$, $\text{c}^{\text{th}}_{\text{max}}$ are the minimum and maximum occlusion risk threshold and $\text{v}^{\text{occ}}_{\text{min}}$, $\text{v}^{\text{occ}}_{\text{max}}$ are the minimum and maximum speed limits when the occlusion risk is sufficient. $v^{occ}_{speed \textunderscore limit}$ has a simple linear relation: it is low when the occlusion risk is high and vice versa. $\text{v}^{\text{occ}}_{\text{min}}$ is for features of expert drivers who take risk and preventing the ego vehicle from freezing and never reaching the goal. If $\text{c}^{\text{th}}_{\text{min}} = 0$ and $\text{v}^{\text{occ}}_{\text{min}} = \text{v}^{\text{occ}}_{\text{max}} = 0$, then our proposed method obtains similar results as over-approximating methods.
\subsubsection{Planning}
Piecewise-jerk speed optimization (PJSO) method is chosen \cite{zhou2020autonomous} for the velocity planning, which involves minimizing the cost function comprising the cost of ride discomfort (acceleration, jerk). The speed limit was added as a hard constraint to the optimization problem:
\begin{equation}
\begin{split}
    \dot{x}(t) < min(\sqrt{a_{lateral \textunderscore max} / \kappa(x)_{max}},\ v^{occ}_{speed \textunderscore limit} )
\end{split}
\end{equation}

where $\kappa(x)_{max}$ is the maximum curvature of the ego vehicle and $a_{lateral \textunderscore max}$ is maximum lateral acceleration along the fixed path. Together $\sqrt{a_{lateral \textunderscore max} / \kappa(x)_{max}}$ represents the approximate speed limit due to the curvature of the fixed path.

\section{Evaluation} \label{evaluation}
\begin{figure}[t!]
    \centering
    \includegraphics[width = 0.75\linewidth]{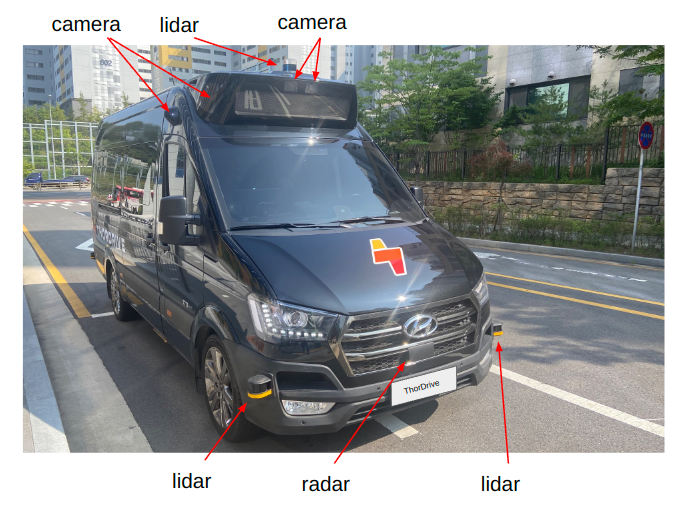}
    \captionsetup{size=footnotesize}
    \caption{The test vehicle. It is remodeled from a Hyundai Solati and equipped with several LIDAR, RADAR, and camera units.}
    \label{fig:sejong_solati}
\end{figure}

\begin{figure}[t]
    \centering
    \begin{subfigure}[b]{0.8\linewidth}        %% or \columnwidth
        \centering
        \includegraphics[width=\linewidth]{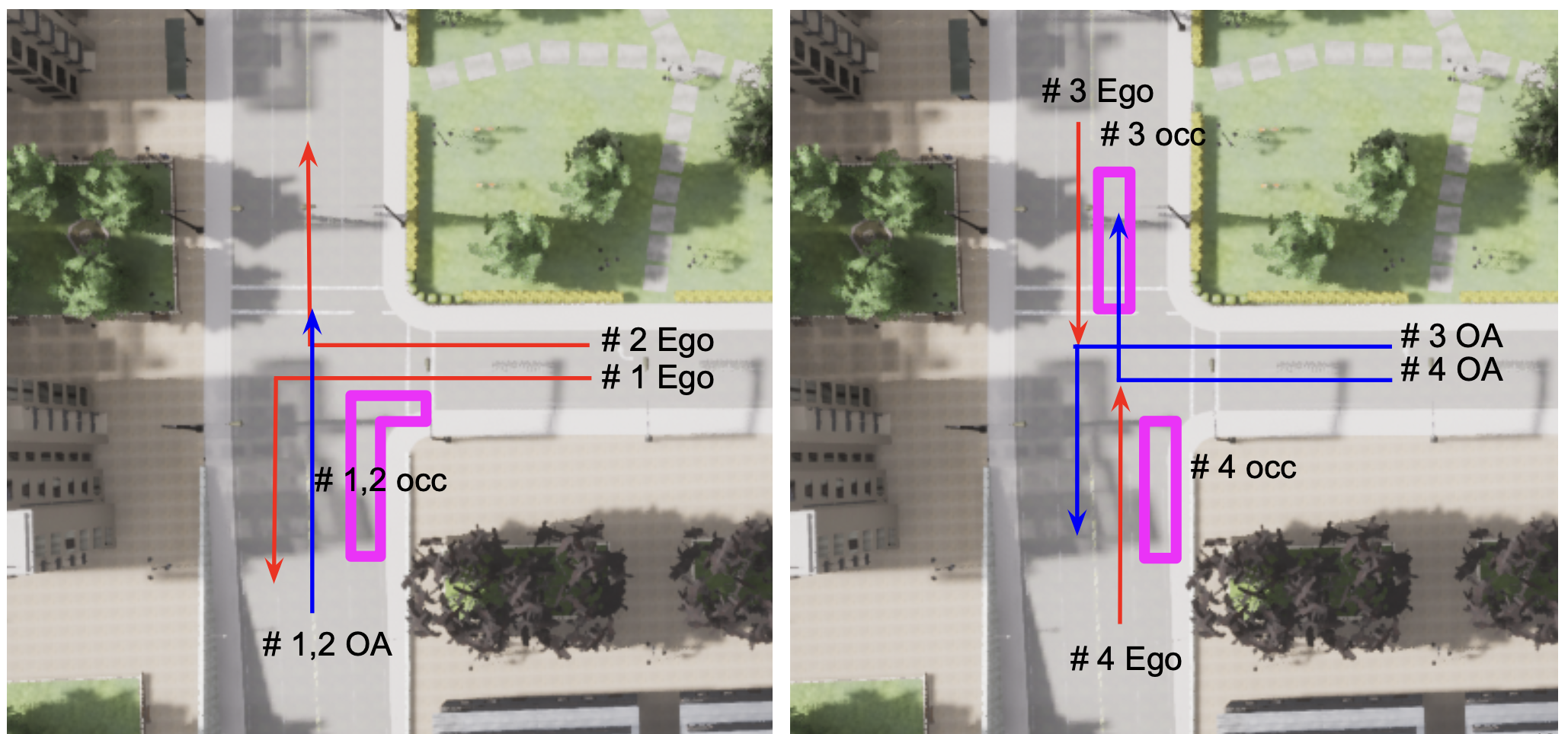}
        \label{fig:evaluation_scenario_a}
    \end{subfigure}
    \begin{subfigure}[b]{0.8\linewidth}        %% or \columnwidth
        \centering
        \includegraphics[width=\linewidth]{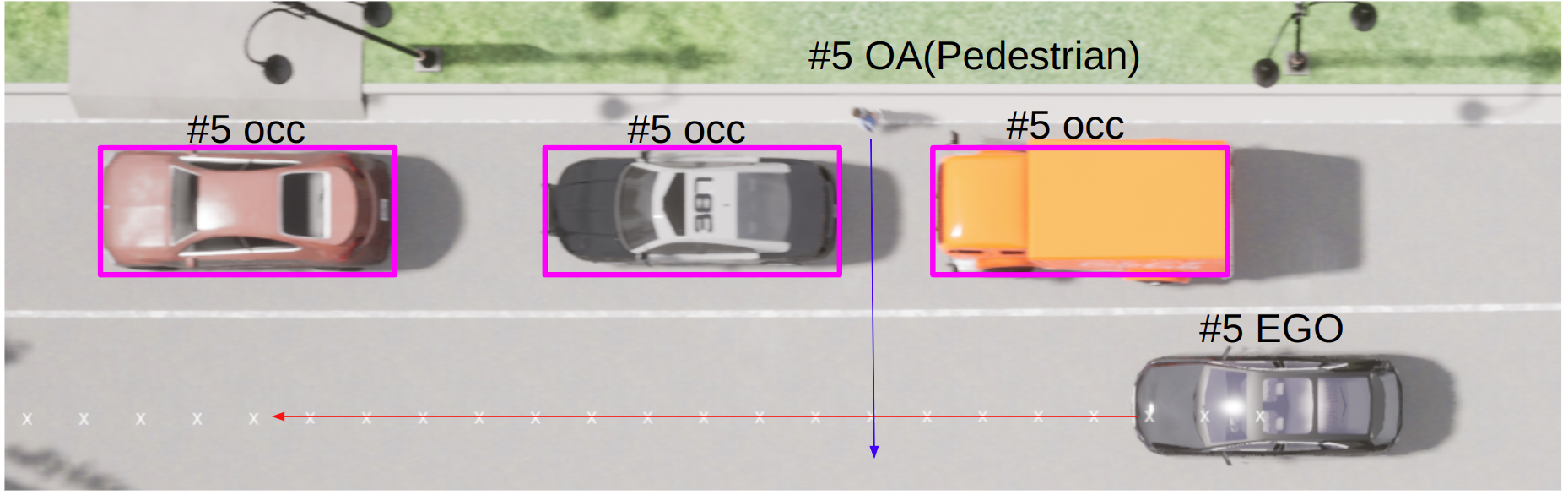}
        \label{fig:evaluation_scenario_b}
    \end{subfigure}
    \captionsetup{size=footnotesize}
    \caption{Five scenarios in the CARLA simulator. Obstacles that induce occlusion are represented as magenta polygon. The ego vehicle drives along the route (red arrow) and counters occluded agent(OA) moving along its own route (blue arrow).}
    \label{fig:evaluation_scenario}
\end{figure}

% To show driving comfort, safety, and computation time, got improved by the proposed method, 
We compared our method with three \emph{baseline} methods in various scenarios in the CARLA simulator and the real world with the test vehicle in Fig. \ref{fig:sejong_solati}. 
% For practical use and computational efficiency, we implemented every baseline and our method to only assess the risk that intersects with the route of the ego vehicle. 
Fig. \ref{fig:evaluation_scenario} shows the five scenarios implemented in CARLA. Furthermore, scenarios 1 and 5 were evaluated in the \emph{real world}. In CARLA, each scenario was simulated 500 times with velocities randomly distributed in $[\text{v}^{\text{road}}_{\text{speed \textunderscore limit}},\ 1.5 \times \text{v}^{\text{road}}_{\text{speed \textunderscore limit}}]$ and pedestrian with velocities randomly distributed in $[\text{4km/h}, \text{6km/h}]$. Every method was implemented using the AMD Ryzen 7 series clocked at 2.2 GHz.

% \begin{figure}[t]
%     \centering
%     \includegraphics[scale=0.22]{evaluation_scenario}
%     \caption{Illustration of scenarios in CARLA simulator. Obstacles that induce occlusion is represented as magenta polygon. The ego-vehicle drives along the route(red) and come across occluded agents(OA) driving along its route(blue).}
%     \label{fig:evaluation_scenario}
% \end{figure}
% \begin{figure}[h]
%     \centering
%     \begin{subfigure}[b]{1.0\linewidth}        %% or \columnwidth
%         \centering
%         \includegraphics[width=\linewidth]{pic/4.IPF_iter_all.pdf}
%         \caption{Resulting paths for each iteration}
%         \label{fig:IPF_each_iterations}
%     \end{subfigure}
%     \\[1ex]
%     \begin{subfigure}[b]{1.0\linewidth}        %% or \columnwidth
%         \centering
%         \includegraphics[width=\linewidth]{pic/4.IPF_iter0.pdf}
%         \caption{Resulting path of the first iteration}
%         \label{fig:IPF_iter0}
%     \end{subfigure}
%     \\[1ex]
%     \begin{subfigure}[b]{1.0\linewidth}        %% or \columnwidth
%         \centering
%         \includegraphics[width=\linewidth]{pic/4.IPF_iter4.pdf}
%         \caption{Resulting path of the last iteration}
%         \label{fig:IPF_iter4}
%     \end{subfigure}
%     \caption{IPF: analysis of each iteration. Blue dots are flattening points $\Omega^{flat}$ and orange discs are collision checking range of the B-spline curve.}
%     \label{fig:IPF_iteration_analysis}
% \end{figure}
% \begin{figure}[h!]
%     \centering
%     \includegraphics[scale=1, height=5cm, width=\linewidth]{pic/4.Refinement_acc_1.0.png}
%     \caption{Trajectory Refinement}
%     \label{fig:refinement_analysis}
% \end{figure}
\subsection{Baseline Methods}
\subsubsection{Baseline 1} \emph{Baseline 1} used the \emph{path velocity decomposition} method for trajectory planning without occlusion-aware risk assessment. The path is given by the route of the ego vehicle, and the velocity profile was generated using PJSO.
\subsubsection{Baseline 2} \emph{Baseline 2} was the same as \emph{baseline 1} but with the SOTA occlusion-aware risk assessment algorithm proposed by \cite{yu2019occlusion}. 
Compare to \emph{baseline 2}, our method was more robust in various scenarios and had better computational efficiency. However, the  obtained occlusion risk for PVs was similar to our method, and \cite{yu2019occlusion} cannot be used in scenarios including occluded pedestrians. Thus, we could only compare the computational times of methods for scenarios without occluded pedestrians.
% The comparison of computational times was fair because the same planning strategy (i.e., baseline 1) was used for baseline 2 and our method.
\subsubsection{Baseline 3} \emph{Baseline 3} also used the same planning method (i.e., baseline 1) but with another SOTA occlusion-aware risk assessment algorithm proposed by \cite{koschi2020set}.

% //
% We have used ratio of number of \emph{particles} that collides with the route path and total number of \emph{particles} with weight to assess occlusion risk. Compare to \emph{baseline 2}, our method is generally applicable in various scenarios(including where sudden appearing pedestrians exist) and computationally efficient. Yet the result of assessing occlusion risk of PV is similar. So we use \emph{baseline 2} to compare computational efficiency.
% \subsection{Metrics} \label{metrics}
% //
\subsection{Metrics} \label{metrics}
We used five metrics to evaluate the methods: the discomfort score, collision rate, traversal time, freeze rate, and average computational time. The discomfort score from \cite{yu2019occlusion} was used to represent the discomfort of passengers in the ego vehicle and $\text{a}_{\text{th}}$ was chosen as $\text{4m/s}^{\text{2}}$:
% There are trade-off relationship between collision rate and traversal time. 
% Also there is inverse proportional relationship between discomfort score and collision rate. Because if the ego vehicle wasn't decelerate enough and sudden breaking happens, the ego vehicle would collide before it earns enough discomfort score.
 % \subsubsection{Discomfort score} We use the discomfort score defined in \cite{yu2019occlusion} to represent the discomfort of passengers in the ego-vehicle.
\begin{equation*} \label{eq:discomfort_score}
\begin{split}
    \mbox{Discomfort Score} = {1\over T} \int^T_0 max(0,|a_{ego}(t)| -\text{a}_{\text{th}})dt
\end{split}
\end{equation*}

% The collision rate was simply defined as the number of simulations with collisions divided by the total number of simulations:
% % \subsubsection{Collision Rate} Collision rate is defined simply the number of simulations with collision divided by the total number of simulations.
% \begin{equation*} \label{eq:collision_rate}
% \begin{split}
%     \mbox{Collision Rate} = {\mbox{\# of simulations with collision} \over \mbox{Total \# of simulations}} \times 100 \%
% \end{split}
% \end{equation*}

The freeze is defined as a situation when the ego vehicle doesn't go any further because the occlusion risk presented in front of the ego vehicle prevents the ego vehicle from moving forward.
% \begin{equation*} \label{eq:freeze_rate}
% \begin{split}
%     \mbox{Freeze Rate} = {\mbox{\# of simulations with freeze} \over \mbox{Total \# of simulations}} \times 100 \%
% \end{split}
% \end{equation*}
% The freeze is defined as a situation when the ego vehicle freezes because the occlusion risk presented in front of the ego vehicle is too high.
% however they tend to make the ego vehicle behave conservatively and even freeze in some corner cases where the ego vehicle must take a risk to pass through.

%  Traversal time
 
\begin{figure} [t]
    \centering
    \begin{subfigure}[b]{0.48\linewidth}        %% or \columnwidth
        \centering
        \includegraphics[width=\linewidth]{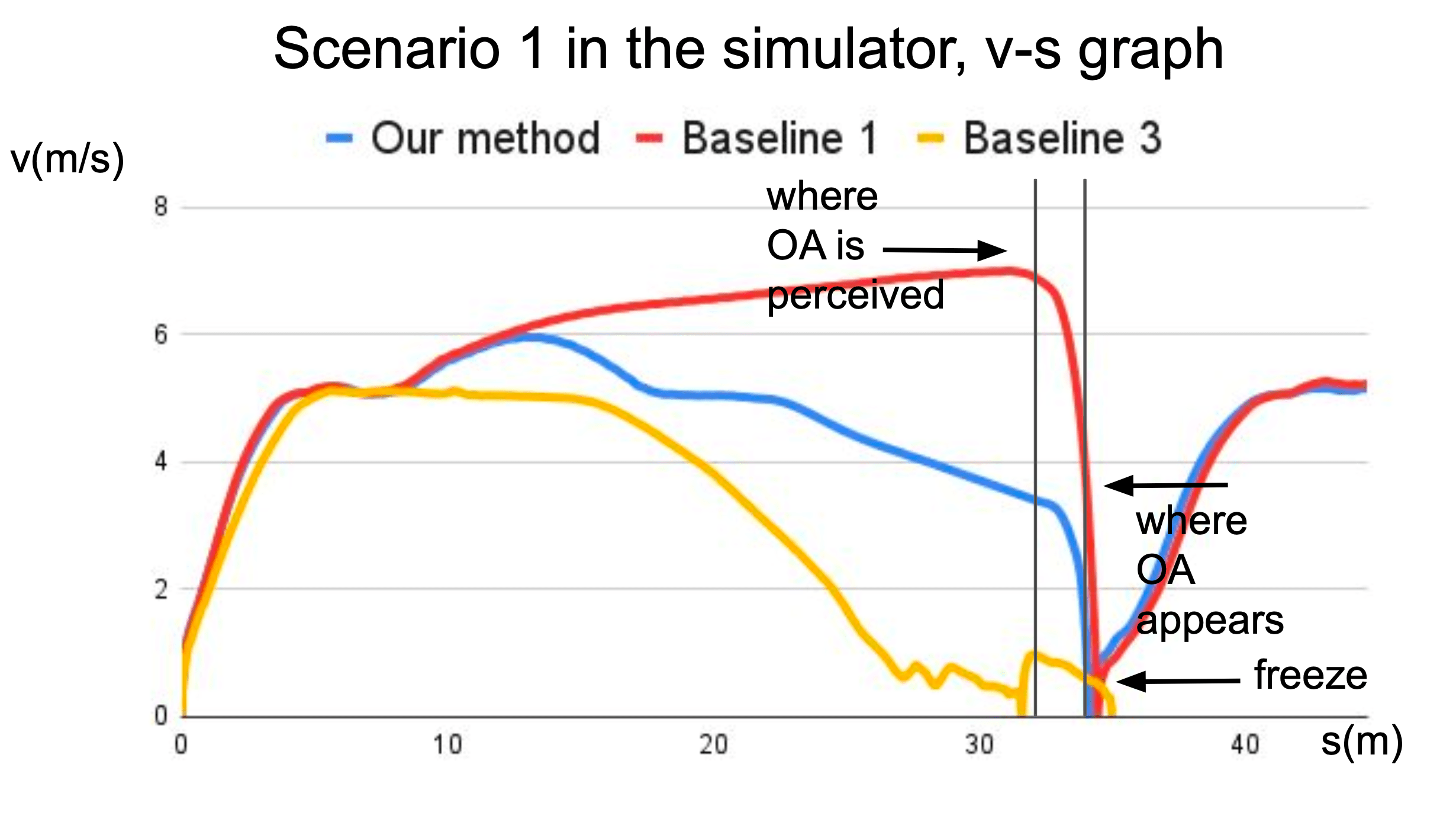}
        \caption{}
        \label{fig:vt_graph}
    \end{subfigure}
    \begin{subfigure}[b]{0.48\linewidth}        %% or \columnwidth
        \centering
        \includegraphics[width=\linewidth]{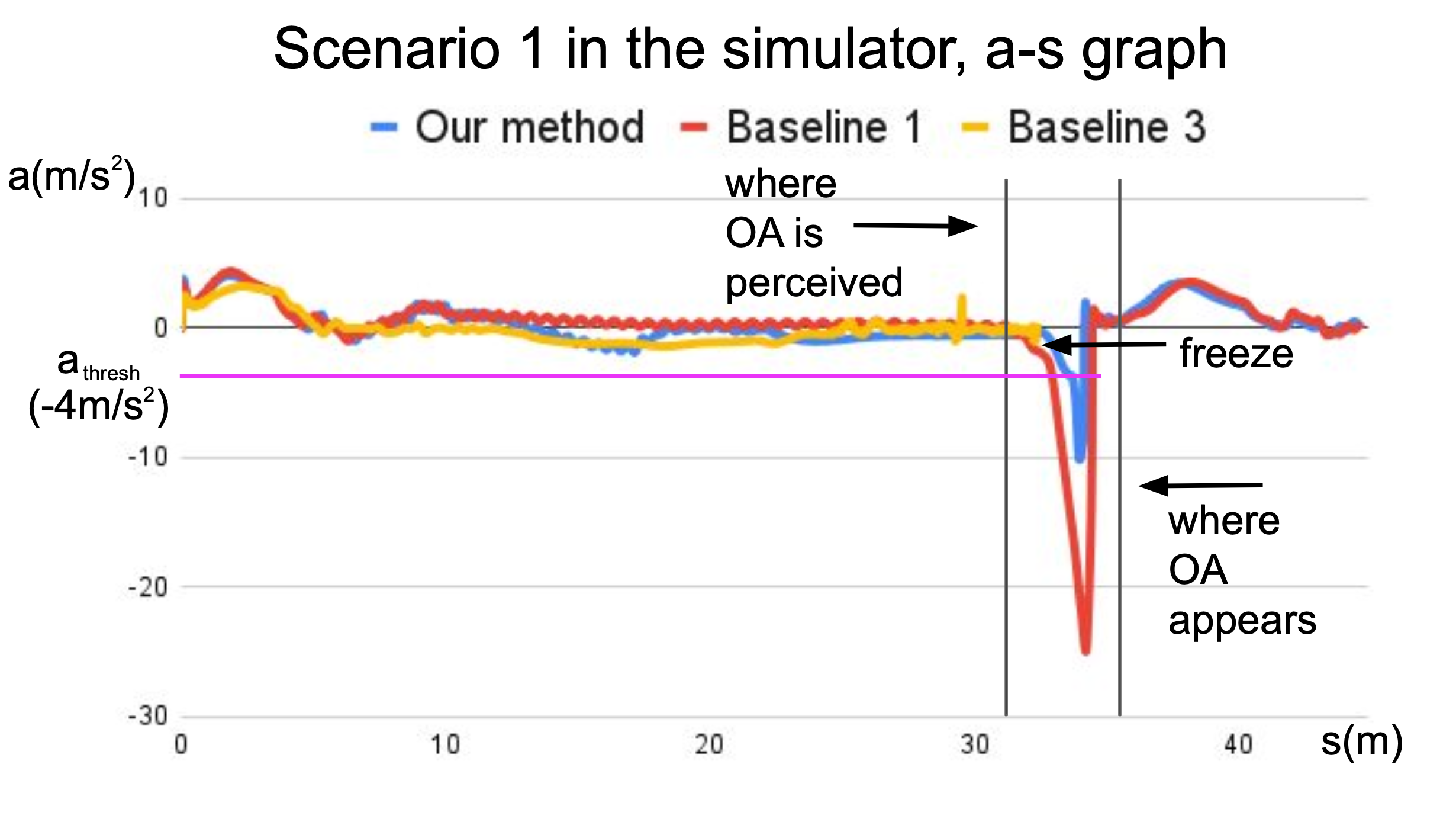}
        \caption{}
        \label{fig:at_graph}
    \end{subfigure}
    \captionsetup{size=footnotesize}
    \caption{Comparison between the proposed method (blue), baseline 1 (red), and baseline 3 (yellow) in the CARLA simulator (Scenario 1): The horizontal axis of graph (i.e., $s$) denotes the longitudinal position of the Frenet frame of the ego vehicle. $\text{a}_\text{th}(\text{4m/s}^\text{2})$ is represented as magenta.}
    \label{fig:vs_as_graph_1}
\end{figure}
\begin{figure} [t]
    \centering
    \begin{subfigure}[b]{0.48\linewidth}        %% or \columnwidth
        \centering
        \includegraphics[width=\linewidth]{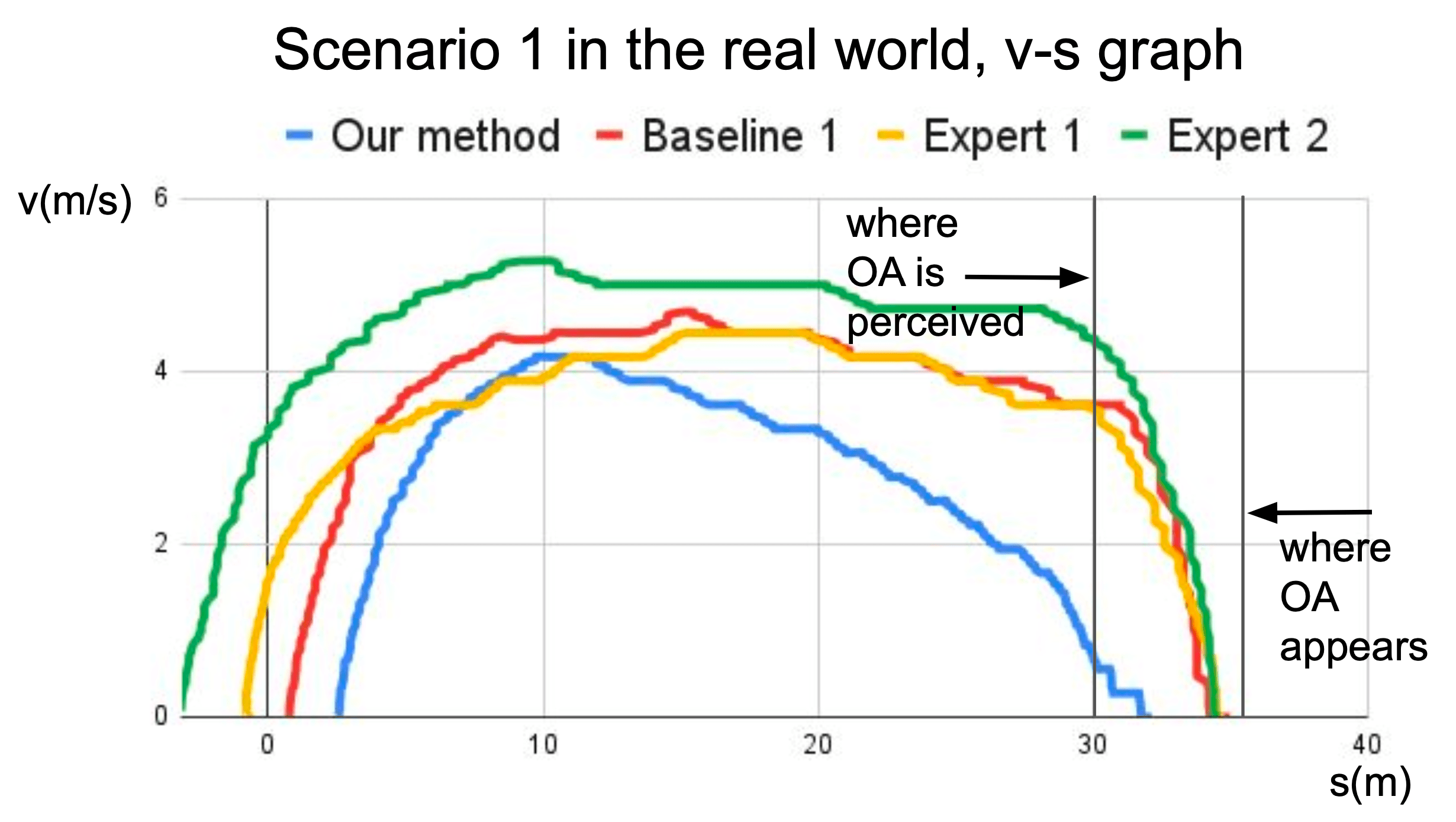}
        \caption{}
        \label{fig:realworld_vs_1}
    \end{subfigure}
    \begin{subfigure}[b]{0.48\linewidth}        %% or \columnwidth
        \centering
        \includegraphics[width=\linewidth]{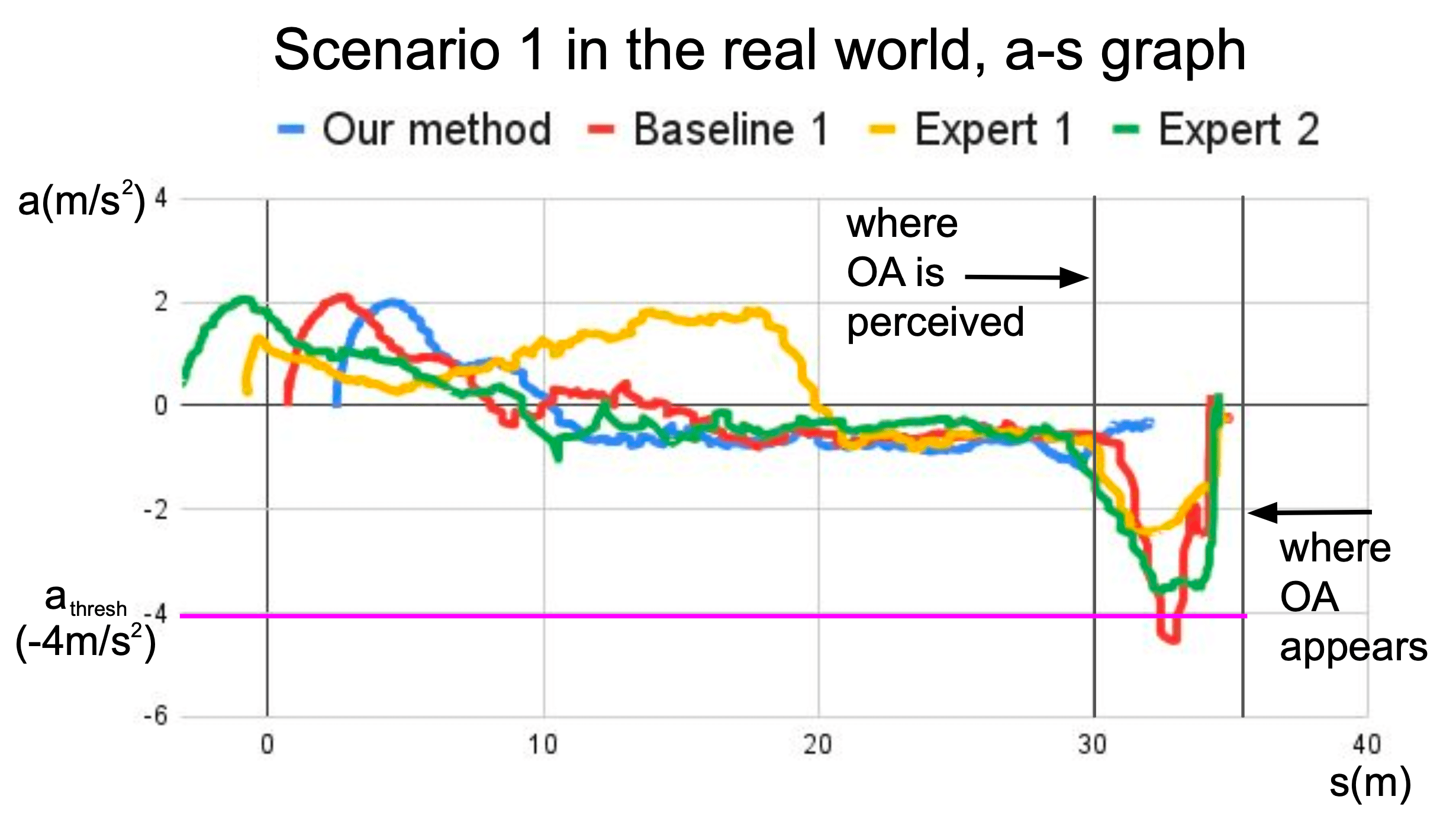}
        \caption{}
        \label{fig:realworld_as_1}
    \end{subfigure}
    \begin{subfigure}[b]{0.48\linewidth}        %% or \columnwidth
        \centering
        \includegraphics[width=\linewidth]{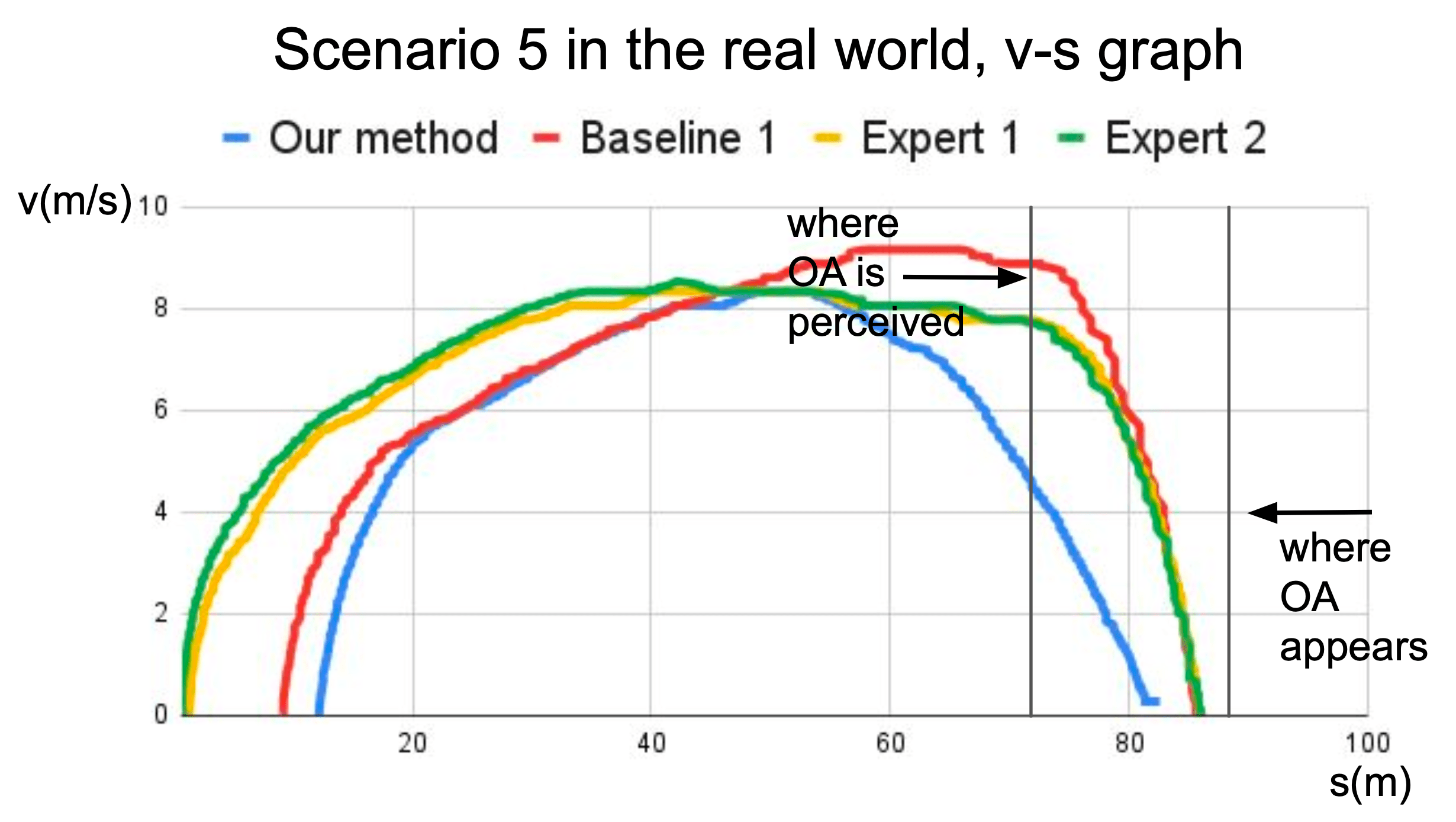}
        \caption{}
        \label{fig:realworld_vs_5}
    \end{subfigure}
    \begin{subfigure}[b]{0.48\linewidth}        %% or \columnwidth
        \centering
        \includegraphics[width=\linewidth]{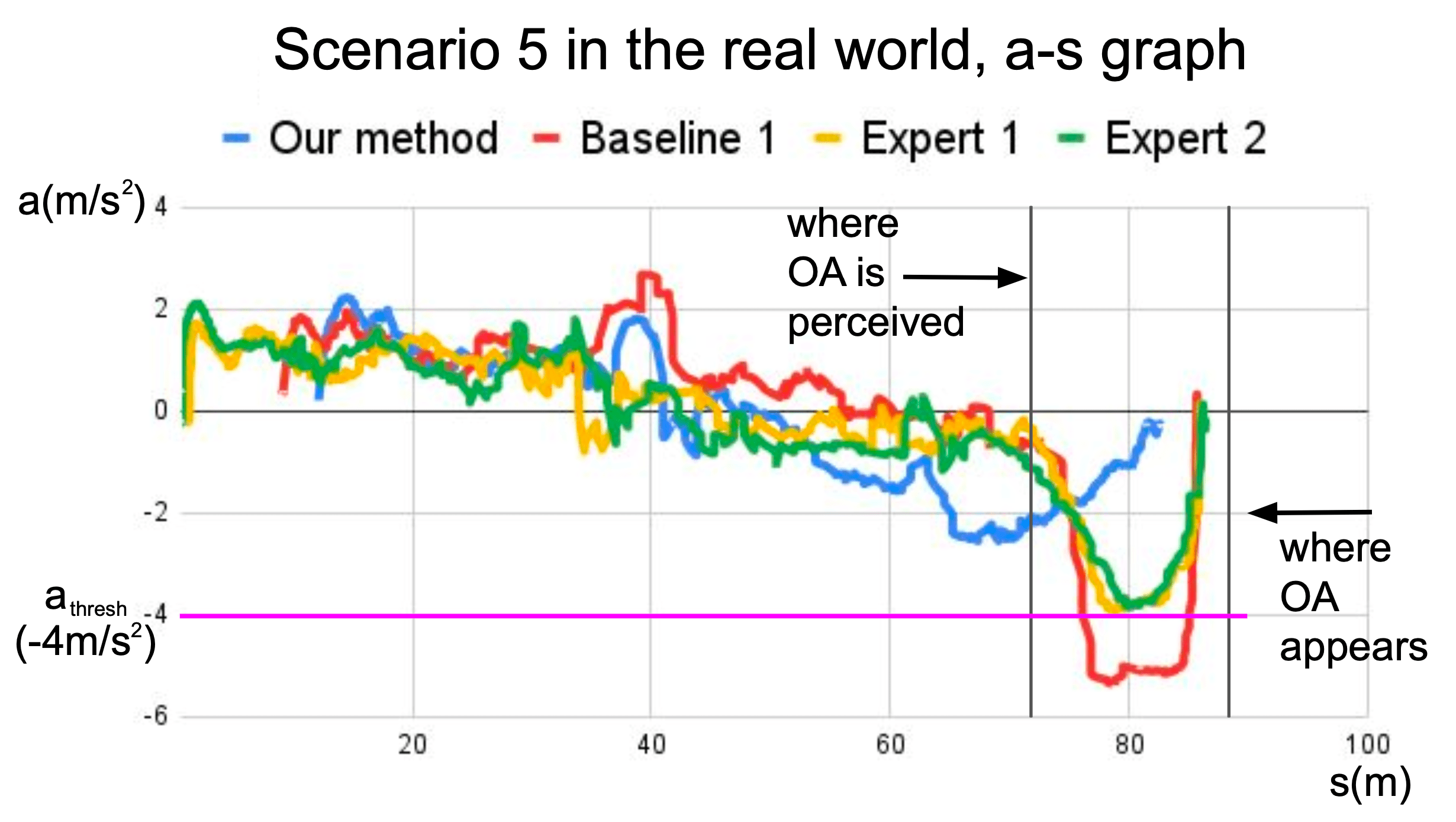}
        \caption{}
        \label{fig:realworld_as_5}
    \end{subfigure}
    \captionsetup{size=footnotesize}
    \caption{Comparison between the proposed method (blue), baseline 1 (red), and expert drivers (yellow and green) in the real world (Scenario 1, 5): The horizontal axis of graph (i.e., $s$) denotes the longitudinal position of the Frenet frame of the ego vehicle. $\text{a}_\text{th}(\text{4m/s}^\text{2})$ is represented as magenta.}
    \label{fig:vs_as_graph_2}
\end{figure}

\section{Result} \label{result}

% Our proposed algorithm showed 'Collision Rate' decreased up to 6.14$\times$, 'Discomfort Score' decreased up to 5.03$\times$, and the lowest increase in 'Traversal Time' is 1.48$\times$ compared to \emph{baseline 1}. The overall result is presented in Table.\ref{table:1}. In Table.\ref{table:1}, scenario 3,4 has resulted better than scenario 1,2. The reason is that the route path of the ego vehicle in scenario 1,2 includes left/right turn, which makes the ego vehicle slow down before the speed limit due to occlusion. Therefore difference of velocity profile between w/ and w/o our method becomes unclear. In Fig.\ref{fig:vt_at_graph} v-t and a-t(deceleration) graphs of an episode of scenario 1 in the simulator are shown. \emph{Baseline 1} and our method both managed to safely stop and avoid collision with a suddenly appearing vehicle. However, our method has reduced the velocity before entering the intersection and prevented itself from sudden deceleration and collision.\\
Compared with baseline 1, our proposed method decreased the collision rate and the discomfort score by up to 6.14 times and 5.03 times, respectively, while it increased the traversal time by 1.48 times. In addition, our method decreased traversal time by up to 1.58 times, while it increased the collision rate by 0.4\%, compared to baseline 3. In Table.\ref{table:1}, baseline 1 showed the shortest traversal time, while resulting in the highest collision rate and discomfort score. This is because baseline 1 doesn't consider the sudden appearance of OAs. On the other hand, baseline 3 was overcautious about OAs, resulting in the longest traversal time or even freeze in heavily occluded scenarios, while having the lowest collision rate and discomfort score. However, the proposed method efficiently assessed occlusion risk, resulting in a small increase in collision rate and discomfort score while decreasing considerable traversal time. The proposed method performed better in scenarios 3 and 4 than in scenarios 1 and 2. This is because the routes of the ego vehicle in scenarios 1 and 2 included left/right turns, which made the ego vehicle slow down before reaching the speed limit due to occlusion. Therefore, the differences between the velocity profiles with our method and with baseline 1 were unclear.

Fig. \ref{fig:vs_as_graph_1} compared the proposed method, baseline 1, and baseline 3 for scenario 1 in the CARLA simulator and Fig. \ref{fig:vs_as_graph_2} compared the proposed method, baseline 1, and expert drivers for scenarios 1 and 5 in the real world. In Fig.\ref{fig:vs_as_graph_1}, every method managed to safely stop and avoid collision with a suddenly appearing agent. However, our method reduced the velocity before entering the intersection which prevented sudden deceleration and collision while baseline 1 didn't reduce the velocity and ended up with sudden deceleration. In addition, baseline 3 was overcautious about OAs and reduced the velocity more than our method before entering the intersection. At the intersection, baseline 3 froze because of its conservativeness. In Fig.\ref{fig:vs_as_graph_2}, expert drivers showed safe and comfort driving without losing efficiency. Our method reduced the velocity earlier than expert drivers, which made the driving much comfortable but lost some efficiency. This trade-off between driving comfort and efficiency is inevitable. Our method can adjust the magnitude of the trade-off by tuning the parameters such as threshold distance and $\text{v}^{\text{occ}}_{\text{min}}$. A video of the experiments is available at \url{https://youtu.be/TJo2pfhkxw4}.\\

\begin{table}[t!]
\centering
\captionsetup{size=footnotesize}
\caption{Performance Comparison Between the Proposed Method, Baseline 1, and Baseline 3 in the CARLA simulator}
\begin{tabular}
{ |p{0.2cm} p{1.2cm}||p{1.0cm}|p{1.2cm}|p{1.0cm}|p{0.8cm}|}
 \hline
 Scenario & & Collision& Discomfort& Traversal& Freeze\\
 & &rate&score&time&rate\\
 \hline
 1& Proposed& 6.00\% & 0.0251& 20.63s& 0.00\%\\
  & Baseline 1& 15.80\% & 0.0663& 14.22s& 0.00\%\\
  & Baseline 3& 0.00\% & 0,0237& 31.54s& 70.00\%\\
 \hline
 2& Proposed& 0.20\%& 0.0008& 16.82s& 0.00\%\\
  & Baseline 1& 0.60\%& 0.0009& 15.00s& 0.00\%\\
  & Baseline 3& 0.00\%& 0.0008& 27.31s& 0.60\%\\
 \hline
 3& Proposed& \textbf{1.40\%}& \textbf{0.0009}& \textbf{12.97s}& 0.00\%\\
  & Baseline 1& \textbf{8.60\%}& \textbf{0.0047}& \textbf{8.75s}& 0.00\%\\
  & Baseline 3& 0.00\%& 0.0008& 21.63s& 0.00\%\\
 \hline
 4& Proposed& \textbf{0.40\%}& \textbf{0.0004}& \textbf{17.4s}& 0.00\%\\
  & Baseline 1& 1.20\%& 0.0020& 11.1s& 0.00\%\\
  & Baseline 3& \textbf{0.00\%}& \textbf{0.0003}& \textbf{27.6s}& 0.00\%\\
 \hline
 5& Proposed& 10.00\%& 0.0065& 13.78s& \textbf{0.00\%}\\
  & Baseline 1& 24.00\%& 0.0186& 12.58s& \textbf{0.00\%}\\
  & Baseline 3& 0\%& 0.0012& 30.3s& \textbf{97.80\%}\\
  \hline  
\end{tabular}
% \captionsetup{size=footnotesize}
% \caption{Table of comparing metrics of ORA(Our method) and Baseline1}
\label{table:1}
\end{table}

% \begin{table}[t!]
% \centering
% \captionsetup{size=footnotesize}
% \caption{Performance Comparison Between the Proposed Method and Baseline 1 in the real world}
% \begin{tabular}
% { |p{0.2cm} p{1.2cm}||p{1.5cm}|p{1.5cm}|p{1.5cm}|p{1.5cm}}
%   \hline
%   Scenario & & Collision& Discomfort\\
%  & &rate&score\\
%  \hline
%  1& Baseline 1& 0.00\%& 0.0206\\
%   & ORA& 0.00\%& 0.00\\
%   \hline
%   &cf&-& 2.86 $\times $DEC\\
%   \hline
%   &Expert&0.00\% & 0.00\\
%   \hline
%   5& Baseline 1& 10.00\%& 0.0578\\
%   & ORA& 0.0\%& 0.00\\
%   \hline
%   &cf&2.40 $\times$ DEC& 2.86 $\times $DEC\\
%   \hline
%   &Expert&0.00\% & 0.00\\
%   \hline
% \end{tabular}
% \label{table:2}
% \end{table}

% We have compared the average 'Computation Time' of \emph{baseline 2} with our method in various scenarios. Even though we set \emph{baseline 2} to have a number of particles $N_k \leq 4 \cdot 10^4$ which was much smaller than it was used for evaluation in \cite{yu2019occlusion}, we have reduced 'Computation Time' up to 20.15$\times$ compared to \emph{baseline 2} due to our novel \emph{Simplified Reachability Quantification}. The overall average computation time result is in Table.\ref{table:2}:
Table.\ref{table:2} compares the average computational times of \emph{baseline 2, 3,} and our method. Even though we set \emph{baseline 2} to $\text{N}_\text{k} \leq 4 \cdot 10^4$, which is a much smaller number of particles than used by \cite{yu2019occlusion}, our method reduced computational time by up to 20.15 times. Moreover, our method quantified the risk of PAs while having a similar computational efficiency to baseline 3.
\begin{table}[t!]
\centering
\captionsetup{size=footnotesize}
\caption{Average Computational Times of the Proposed Method, Baseline 2, and Baseline 3 in the CARLA simulator.}
\begin{tabular}
{ |p{1.2cm}||p{1.3cm}|p{1.3cm}|p{1.3cm}|p{1.3cm}|}
 \hline
 \multicolumn{5}{|c|}{Computation Time Table} \\
 \hline
 Scenario \#& 1& 2& 3& 4\\
 \hline
 Proposed& \textbf{1.3029ms}& 1.5005ms& 1.6467ms& 2.6913ms\\
 \hline
 Baseline 2& \textbf{26.2528ms}& 18.6865ms& 32.0241ms& 39.9024ms\\
 \hline
 Baseline 3& \textbf{1.2693ms}& 1.3453ms& 1.4322ms& 2.3604ms\\
 \hline
\end{tabular}
\label{table:2}
\end{table}
% \begin{figure}[h]
%     \centering
%     \includegraphics[scale=0.4]{vt_graph}
%     \caption{v-t graph of baseline 1 and our method in scenario 3}
%     \label{fig:vt_graph}
% \end{figure}

% \begin{figure}[h]
%     \centering
%     \includegraphics[scale=0.4]{at_graph}
%     \caption{a-t graph of baseline 1 and our method in scenario 3}
%     \label{fig:as_graph}
% \end{figure}

% or


\section{Conclusions and Future Work} \label{conclusion_and_future_work}
We proposed the occlusion-aware risk assessment method and planning strategy using SRQ that improves upon the work of other authors \cite{yu2019occlusion,yu2020risk,koschi2020set,orzechowski2018tackling} by extending the risk assessment to occluded pedestrians and reducing the computational load. We evaluated our method in various scenarios and the results showed that our method effectively decreased the collision rate and discomfort score while greatly reducing the computational time compared with the current state-of-the-art methods. Future work will involve using sequential reasoning to remove redundant PAs and improve the driving efficiency of the ego vehicle by preventing conservative behavior.

\ifCLASSOPTIONcaptionsoff
  \newpage
\fi

\bibliographystyle{unsrt}
\bibliography{references}

\end{document}